%% file: main.tex
\documentclass{article}

\usepackage{microtype}
\usepackage{graphicx}
\usepackage{booktabs} 
\usepackage{float}
\usepackage{svg}
\usepackage{times}
\usepackage{epsfig}
\usepackage{graphicx}
\usepackage{amsmath}
\usepackage{amsthm}
\usepackage{caption}
\usepackage{amssymb}
\usepackage{booktabs}
\usepackage{subfig}
\usepackage[pagebackref=true,breaklinks=true,colorlinks,bookmarks=false]{hyperref}
\usepackage[accepted]{icml2019}
\usepackage{dsfont}

\mathchardef\mhyphen="2D %
\newcommand{\DPN}{DoPaNet}

\newtheorem{theorem}{Theorem}[section]
\newtheorem{corollary}{Corollary}[theorem]
\newtheorem{lemma}[theorem]{Lemma}

\newtheorem*{lemma1}{Lemma 3.1}
\newtheorem*{theorem2}{Theorem 3.2}
\newtheorem*{corollary3}{Corollary 3.2.1}

\usepackage{etoolbox}
\makeatletter
\patchcmd\@combinedblfloats{\box\@outputbox}{\unvbox\@outputbox}{}{%
   \errmessage{\noexpand\@combinedblfloats could not be patched}%
}%
 \makeatother

\begin{document}
\twocolumn[
\icmltitle{Domain Partitioning Network}

{\centerline{
\begin{tabular}[t]{c}
Botos Csaba$^{1, 2}$,\; 
Adnane Boukhayma$^1$,\; 
Viveka Kulharia$^1$, 
Andr\'{a}s Horv\'{a}th$^2$,\; 
Philip H. S. Torr$^1$\\
$^1$ \textnormal{University of Oxford, United Kingdom}\\
$^2$ \textnormal{P\'{a}zm\'{a}ny P\'{e}ter Catholic University, Hungary}\\
{\small \texttt{\{csbotos,viveka\}@robots.ox.ac.uk,\{adnane.boukhayma,philip.torr\}@eng.ox.ac.uk}}\\
{\small \texttt{horvath.andras@itk.ppke.hu}}
\end{tabular}\hspace{1cm}
}}

\icmlkeywords{generative adversarial networks, multi-adversarial network}

\vskip 0.3in
]

\begin{abstract}
    Standard adversarial training involves two agents, namely a generator and a discriminator, playing a mini-max game. However, even if the players converge to an equilibrium, the generator may only recover a part of the target data distribution, in a situation commonly referred to as \textit{mode collapse}. In this work, we present the Domain Partitioning Network (\DPN{}), a new approach to deal with mode collapse in generative adversarial learning. We employ multiple discriminators, each encouraging the generator to cover a different part of the target distribution. To ensure these parts do not overlap and collapse into the same mode, we add a classifier as a third agent in the game. The classifier decides which discriminator the generator is trained against for each sample. Through experiments on toy examples and real images, we show the merits of \DPN{} in covering the real distribution and its superiority with respect to the competing methods. Besides, we also show that we can control the modes from which samples are generated using \DPN{}.
\end{abstract}

\begin{figure}[th!]
    \centering
    \includegraphics[width=0.4\textwidth]{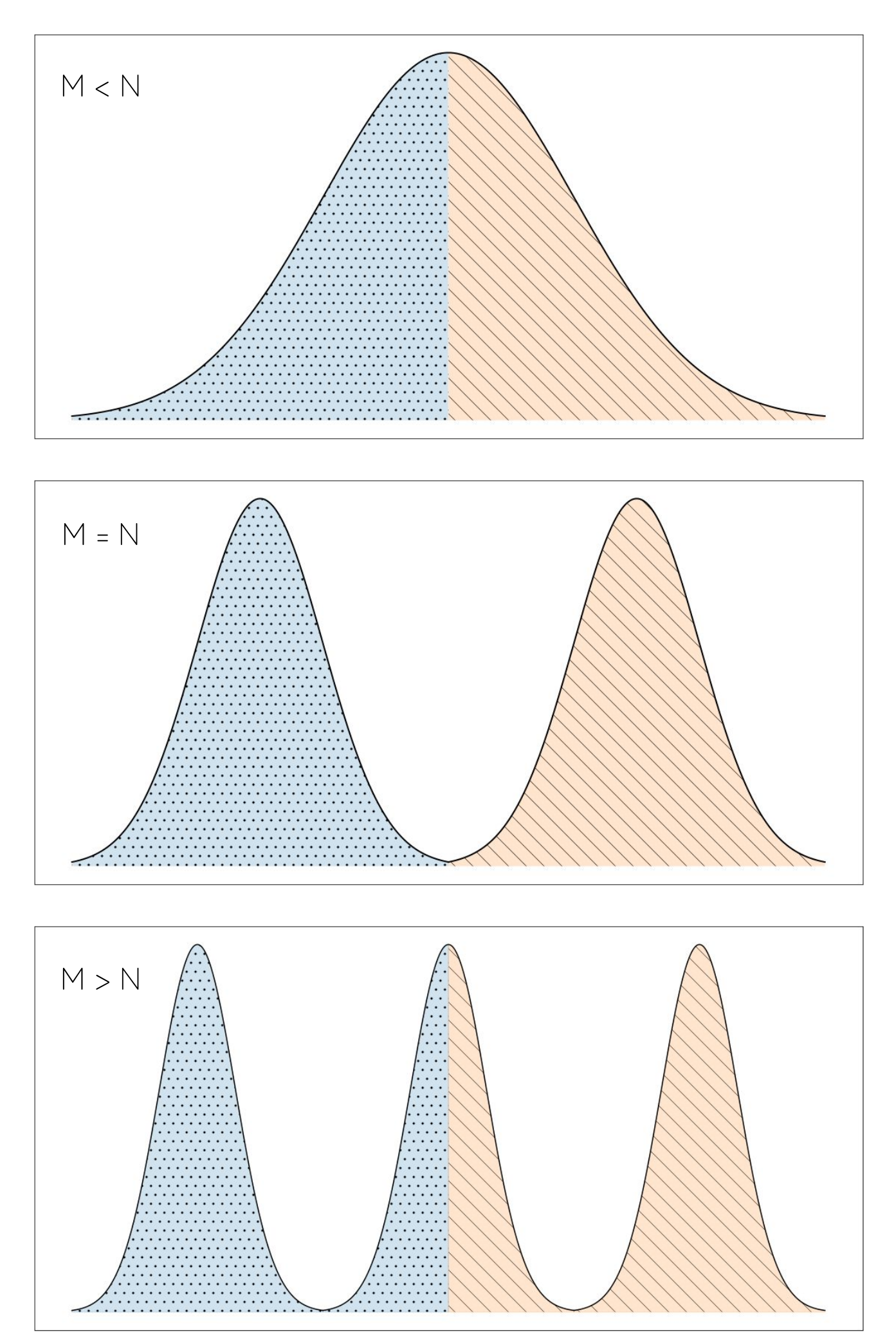}
    \caption{
        Illustration of the expected behaviour \DPN{} using two discriminators ($N=2$), in case of a Uni-modal ($M=1$, top), bi-modal ($M=2$, middle) and tri-modal ($M=3$, bottom) target distribution.
        The classifier ensures that the generated modes (in orange and blue) corresponding to two different discriminators do not overlap.
    }
    \label{fig:1}
    \vspace{-10pt}
\end{figure}

\section{Introduction}
Generative Adversarial Networks \cite{Goodfellow14} (GANs) consist of a deep generative model which is trained through a minimax game involving a competing generator and discriminator. The discriminator is tasked to differentiate real from fake samples, whereas the generator strives to maximize the mistakes of the discriminator. At convergence, the generator can sample from an estimate of the underlying real data distribution. The generated images, are observed to be of higher quality than models trained using maximum likelihood optimization. Consequently, GANs have demonstrated impressive results in various domains such as image generation \cite{gulrajani2017improved}, video generation \cite{vondrick2016generating}, super-resolution \cite{ledig2017photo}, semi-supervised learning \cite{donahue2016adversarial} and domain adaptation \cite{zhu2017unpaired}.

\begin{figure}[t!]
    \centering
    \includegraphics[width=0.45\textwidth]{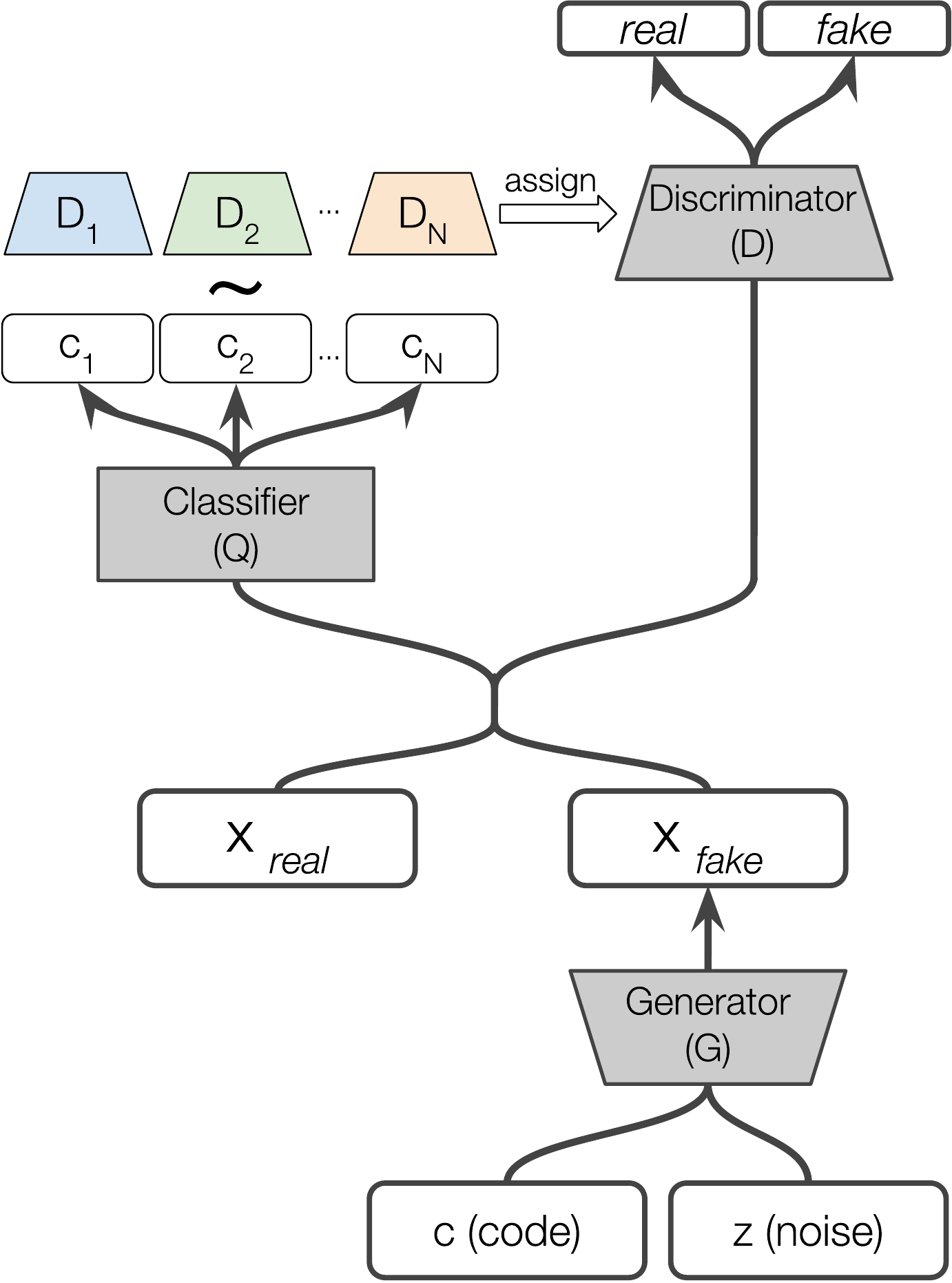}
    \caption{\DPN{}, our proposed framework.
    Here $\sim$ denotes a sampling operation from the categorical probability distribution $Q(x)$. $c$ is a categorical code with one-hot encoding.
    Using the resulting category index $\sigma\sim Q(x)$ we select the corresponding discriminator $D_{i=\sigma}$ and connect it to the computation graph.
    From the perspective of the real sample $x$ and the generated sample $\hat{x}$ the respective computation graph is fully-differentiable, and can be trained like the standard GAN \cite{goodfellow2014generative}. 
    }
    \label{fig:routernet}
    \vspace{-10pt}
\end{figure}

GANs are trained with the objective of reaching a Nash-equilibrium \cite{mescheder2018convergence}, which refers to the state where neither the discriminator nor the generator can further enhance their utilities unilaterally. However, the generator might miss some modes of the distribution even after reaching the equilibrium as it can simply fool the discriminator by generating from only few modes of the real distribution
\cite{goodfellow2016nips,arjovsky2017towards,che2016mode,chen2016infogan,
salimans2016improved}, and hence producing a limited diversity in samples. To address this problem, the literature explores two main approaches: Improving GAN learning to practically reach a better optimum \cite{arjovsky2017towards, 
metz2016unrolled,salimans2016improved,arjovsky2017wasserstein,
gulrajani2017improved,berthelot2017began}, or explicitly forcing GANs to produce various modes by design \cite{chen2016infogan,ghosh2017multi,durugkar2016generative,
che2016mode,liu2016coupled}. We hereby follow the latter strategy and propose a new way of dealing with GAN mode collapse.
By noticing that using a single discriminator often leads to the generator covering only a part of the data, we bring more discriminators to the game such that each incentivises the generator to cover an additional mode of the data distribution. For each discriminator to focus on a different target mode, we introduce a third player, a classifier $Q$ that decides the discriminator to be trained using a given real sample. To ensure that these various target data modes do not collapse into the same mode, the classifier $Q$ also decides the discriminator to train the generator for a given generated sample. We find that this strategy, illustrated in Figure \ref{fig:routernet}, yields better coverage of the real data distribution at convergence and simultaneously improves the stability of the training as well. 

We showcase our method on demonstrative toy problems and show that it outperforms competing methods in avoiding mode collapse. We show that the $Q$ network is able to distinguish different modes of the real data and therefore each discriminator works on a separate mode. This ensures that the generator can sample from a different mode for every input code vector. We also show \DPN{}'s ability to generate good quality and diversified images covering various modes present in the datasets of real images.  

We also provide theoretical analysis to show that at global optimum of the objective, the generator replicates the real distribution, categorized into different modes such that it can sample from any mode given the corresponding code vector $c$.

\section{Related work}
There is a rich literature on improving training stability and increasing sample diversity for GANs. We only focus on a selection of works that relate closely to ours. 
\cite{pmlr-v70-arora17a} introduces theoretical formulation stating the importance of multiple generators and discriminators in order to completely model the data distribution. GMAN \cite{durugkar2016generative} proposes using multiple discriminators. 
They explore 3 settings where the generator can either be trained against the best discriminator, the averaged discriminators, or the weighted averaged discriminators. This helps training the network without modifying the minimax objective. Even though they use multiple discriminators, all of them are trained using all of the available real data, which does not explicitly help in avoiding mode collapse. We improve on this strategy by adding a classifier as a third component, with the task of choosing the discriminator for the given input sample during training, therefore each of the multiple discriminators specializes on a different part of the real data distribution. We also compare \DPN{} with GMAN \cite{durugkar2016generative} in our experiments (Section \ref{sec:exp}). 
Triple-GAN \cite{li2017triple} incorporates a classifier in the adversarial training but it focuses on semi-supervised learning and therefore it needs some part of the real data to be labeled during training. It uses only one discriminator which is also conditioned on the sample labels. Contrarily our aim is to circumvent the mode collapse problem in the general case where the labels of the samples may not be available.
InfoGAN~\cite{chen2016infogan} uses a $Q$ network to maximize mutual information between the input code to the generator and its generated samples. It helps in disentangling several factors of variation, e.g.\ writing styles in case of digits, pose from lightning, etc. It is different from our approach as it uses the $Q$ network as well to train the generator. Hence it is possible that the generator colludes with $Q$ in disentangling the factors of variation, but simultaneously fooling the discriminator, while sampling from only few modes of the data. It can therefore still face the mode collapse problem which we show in the experiments (Section~\ref{sec:exp}).
Several works propose using multiple generators~\cite{arora2017generalization,ghosh2016message,liu2016coupled}. For instance, MAD-GAN~\cite{ghosh2017multi} improves the learning by compelling the generators to produce diverse modes implicitly using the discriminator. This is achieved by requiring the discriminator to identify the generator that produced the fake samples along with recognizing fake samples from reals. The discriminator does not explicitly force each generator to capture a different mode, while in our case the generator is urged to capture distinct modes by being trained with different discriminators. We also show \DPN{}'s superiority over MAD-GAN in our experiments (Section~\ref{sec:exp}).

\section{Method}
    In this section we first briefly discuss the preliminaries (\ref{subsec:prelim}): the general objective for training Generative Adversarial Nets and conditional sampling and training.
    Then we detail the objective of \DPN{} (\ref{subsec:our}) and how we optimize it.
    
    \subsection{Preliminaries}\label{subsec:prelim}
        \paragraph{Generative Adversarial Networks} 
        Generative adversarial networks can be considered as a game, where players in the form of neural networks are optimized against each other. Let $p_d$ be the $real$ data distribution and $p_g$ be the distribution learnt by the generator $G$.
        Different tasks are assigned to the players: firstly, the generator $G$ takes an input noise $z \sim p(z)$ and returns a sample $\hat{x}=G(z)$.
        The discriminator $D$ takes an input $x$ which can either be a real sample from the training set or a sample produced by the generator. The discriminator then outputs a conditional probability distribution over the source of the sample $x$. In practice $D$ is a binary classifier that ideally outputs $1$ if the sample is real and $0$ if the sample is fake. Formally the following min-max objective is iteratively optimized:

        \begin{equation}\label{eq:ls}
        \begin{split}
                \mathop{\min}_{G}\mathop{\max}_{D}V(D,G) :=& \mathop{\mathbb{E}}_{x \sim p_{d}}\left[\log D \left(x\right)\right]\\
                &+\mathop{\mathbb{E}}_{z \sim p_{z}}\left[\log \left(1-D\left(G\left({z}\right)\right)\right)\right]
        \end{split}
        \end{equation}
        
        The parameters of D are updated to maximize the objective while the generator $G$ is trained to minimize it.
        
        \paragraph{Conditional generation}
        We can condition the modeled distribution by making $G$ take a code vector $c$ as an additional input to produce a sample $\hat{x}_{c} = G(z, c)$, as it is done in InfoGAN~\cite{chen2016infogan} and other conditional variants~\cite{mirza2014conditional}. In our case, we restrict the code vector $c$ to have a one-hot encoding.
        Defining the conditional probability distribution as $Q(x)=p_{c \mid x}$, we obtain an objective function for the classifier $Q$, the general cross-entropy loss:
        \begin{equation}\label{eq:lc}
                \mathop{\min}_{Q}L(Q,G) := \mathop{\mathbb{E}}_{z\sim p_{z}, c\sim p_{c}}\left[CE(c, Q\left(G\left(z, c)\right)\right)\right]
        \end{equation}
        
        where $CE(.,.)$ is the cross entropy function. The conditional variants of the standard GAN settings optimize both Objectives~(\ref{eq:ls}) and~(\ref{eq:lc}), where $G$ may or may not be optimized over Objective~(\ref{eq:lc}). We do not use $G$ to optimize the Objective~(\ref{eq:lc}).

\begin{algorithm*}[t!]
\caption{\DPN{} training algorithm}\label{alg:dpn}
\begin{algorithmic}[1]
\FOR{number of training iterations}
    \STATE{Sample minibatch of $m$ noise samples $\{ {z}^{(1)}, \dots, {z}^{(m)} \}$ from the noise prior $p_z({z})$.}
    \STATE{Sample minibatch of $m$ code samples $\{ {c}^{(1)}, \dots, {c}^{(m)} \}$ from the code prior $p_c({c})$.}
    \STATE{Sample minibatch of $m$ examples $\{ {x}^{(1)}, \dots, {x}^{(m)} \}$ from the data generating distribution $p_d({x})$.}
    \STATE{Update $Q$ by ascending its stochastic gradient:
        \[
            \nabla_{\theta_q} \frac{1}{m} \sum_{i=1}^m  c^{(i)} \cdot \log Q\left(G\left(z^{(i)}, c^{(i)}\right)\right) 
        \]}
    \STATE{Decide for every input which of the $N$ discriminators to use by sampling from the likelihood distribution of $Q$:
        \[
            \sigma (i) \sim Q\left(x^{(i)}\right) 
            \qquad 
            \hat{\sigma} (i) \sim Q\left(G\left(z^{(i)}, c^{(i)}\right)\right)
        \]
    }
    \STATE{$\forall n \in [1, \dots, N]$, define the set of samples that are assigned to the $n^\text{th}$ discriminator $D_n$ as:}
    \[
        \mathcal{D}_n = \left\{x^{(i)} | \sigma (i) = n\right\}
        \qquad
        \hat{\mathcal{D}}_n = \left\{G\left(z^{(i)}, c^{(i)}\right) \rvert \hat{\sigma} (i) = n\right\} 
    \]
    
    \STATE{$\forall n \in [1, \dots, N]$, update the $n^\text{th}$ discriminator by ascending its stochastic gradient:}
    \[
        \nabla_{\theta_{d_n}}
        \left(
        \frac{1}{|\mathcal{D}_n|} \sum_{x \in \mathcal{D}_n}
        \log D_{n}\left(x\right)
        + 
        \frac{1}{|\hat{\mathcal{D}}_n|} \sum_{\hat{x}\in \hat{\mathcal{D}}_n} 
            \log \left(1-D_{n}\left(\hat{x}\right)\right)
        \right)
    \]
    
    \STATE{Sample minibatch of $m$ noise samples $\{ {z}^{(1)}, \dots, {z}^{(m)} \}$ from the noise prior $p_z({z})$.}
    \STATE{Sample minibatch of $m$ code samples $\{ {c}^{(1)}, \dots, {c}^{(m)} \}$ from the code prior $p_c({c})$.}
    \STATE{Decide for every $i^\text{th}$ input which of the $N$ discriminators to use by sampling from likelihood distribution of $Q$:
        \[
            \hat{\sigma} (i) \sim Q\left(G\left(z^{(i)}, c^{(i)}\right)\right)
        \]
    }
    \STATE{Update the generator by descending its stochastic gradient:
        \[
            \nabla_{\theta_g} \frac{1}{m} \sum_{i=1}^m
            \log \left(1-D_{\hat{\sigma} (i)}\left(G\left({z}^{(i)}, {c}^{(i)}\right)\right)\right)
        \]}
        \vspace{-10pt}
\ENDFOR
\end{algorithmic}
\end{algorithm*}

    \subsection{Our approach: \DPN{}}\label{subsec:our}
    \DPN{} consists of three main components: A conditional generator $G$, a classifier $Q$ and a set of independent discriminators $\{D_i\}$.
    We use categorical code vectors $c\in \{0,1\}^N$ with one-hot encoding where $N$ is the number of discriminators used. We use the notation $c_i$ to denote the one-hot code vector $c$ with value at the $i^\text{th}$ index as 1.
    As illustrated in Figure \ref{fig:routernet}, $G$ generates a sample $\hat{x}_c=G(z, c)$.
    Next we feed the sample to the classifier Q to get the categorical probability distribution.
    For each generated sample we draw $\hat{\sigma}\sim Q(\hat{x}_c)$, i.e.\ $\hat{\sigma}\in [1,...,N]$ that decides the corresponding discriminator and $D_{\hat\sigma}$ that is going to process the generated sample. 
    Formally, we define $D(\hat{x}):=D_{\hat\sigma}(\hat{x}_c)$. 
    Similarly, for the real sample $x\sim p_d$, we draw $\sigma \sim Q(x)$ and define the discriminator $D(x):=D_{\sigma}(x)$ for the sample $x$. Thus, for every sample, the discriminator used is decided by the classifier $Q$.
    This yields a fully-differentiable computational graph, despite the fact that the sampling operation $\sigma\sim Q(x)$ is non-differentiable.
    In other words, once $D$ is selected using predictions from $Q$, the training requires no further modifications to the standard GAN optimization algorithm, therefore it is compatible with all recent advanced variants of GANs. 
    In our experiments we define $p_z$ as a standard normal distribution and $p_c$ as a uniform categorical distribution unless otherwise stated.
    
    Let us define the minimax objective for \DPN{}:
    \begin{equation}\label{eq:minmax}
    \begin{split}
        \mathop{\min}_{G}\mathop{\max}_{\{D_i\}_{i=1}^{N}}&M(\{D_i\}_{i=1}^{N},G) := \mathop{\mathbb{E}}_{\substack{x \sim p_{d} \\ \sigma \sim Q(x) }}\left[\log D_{\sigma}\left({x}\right)\right]\\
        &+\mathop{\mathbb{E}}_{\substack{z\sim p_{z}, c\sim p_{c} \\ \hat\sigma \sim Q(G(z, c)) }}\left[\log \left(1-D_{\hat{\sigma}}\left(G\left({z}, {c}\right)\right)\right)\right]\\
    \end{split}
    \end{equation}
    We train \DPN{} by iteratively optimizing the following objective function (refer Algorithm~\ref{alg:dpn}): 
    \begin{equation}
        \mathop{\min}_{G}\mathop{\max}_{\{D_i\}_{i=1}^{N}}M(\{D_i\}_{i=1}^{N},G) + \mathop{\min}_{Q}L(Q,G)
    \end{equation}

\begin{figure*}
    \centering
    \subfloat[Standard GAN]{
        \includegraphics[width=.3\linewidth]{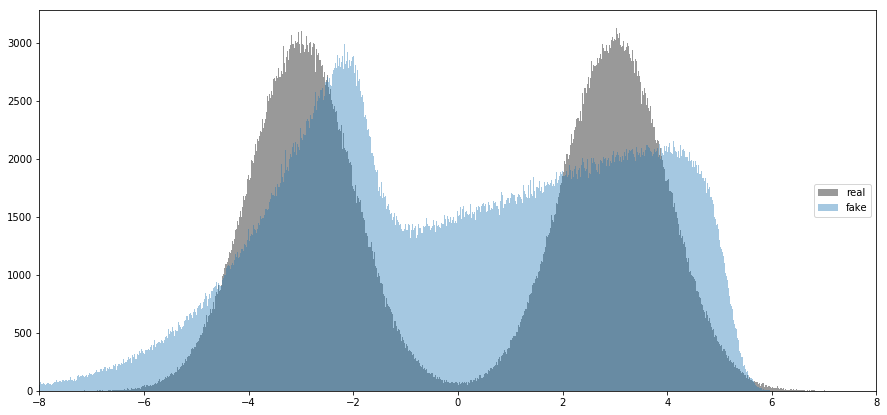}
    }
    \subfloat[\DPN{}, 2 modes, 2 Disc.]{
        \includegraphics[width=.3\linewidth]{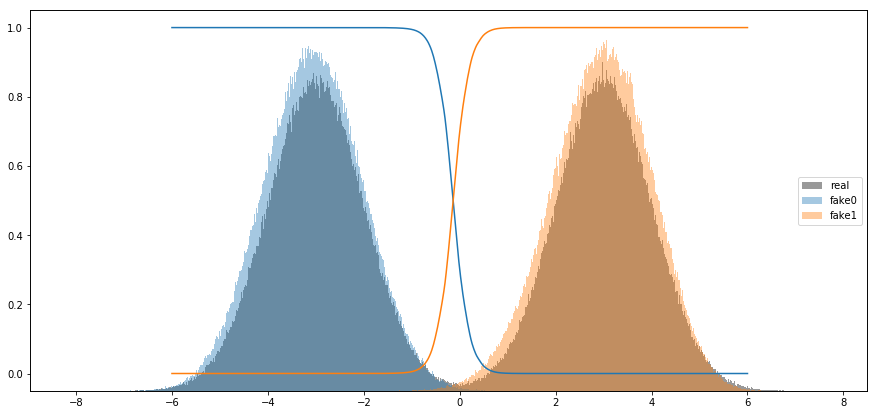}
    }
    \subfloat[\DPN{}, 3 modes, 2 Disc.]{
        \includegraphics[width=.3\linewidth]{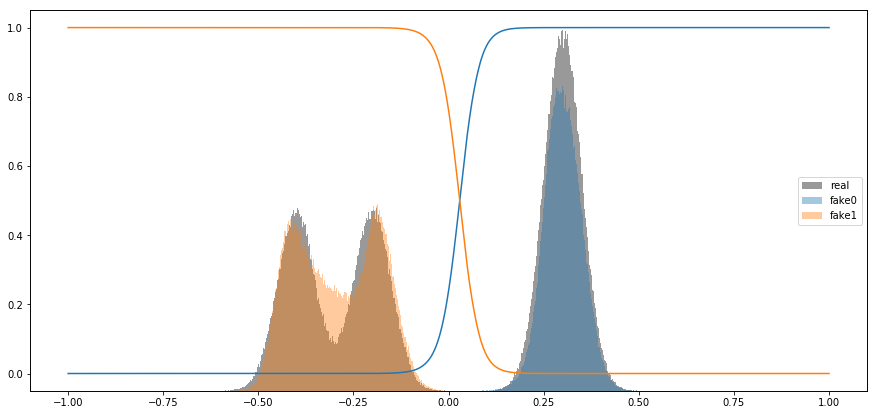}
    }
    \caption{Theoretical analysis: The figures are plotted using 100000 samples and 1000 bins histogram where the grey area represents the real data distribution. \textit{Fig. (a)}: Standard GAN - In special case, $N=1$ \DPN{} is equivalent to the standard GAN. The blue area represents the model distribution. \textit{Fig. (b, c)}:  The orange and blue area represent the generations corresponding to $c_1$ and $c_2$. The number of discriminators is fixed at $N = 2$, while the number of modes is $M = 2, 3$ respectively. The orange and blue curves depict the predicted class probability of $c_1$ and $c_2$ respectively by the classifier $Q$. It can be seen that support of $p_d^1$ (i.e.\ $x<0$) can be considered disjoint from the support of $p_d^2$ (i.e.\ $x>0$) due to steep change in $p_{c \mid x}$ around $x=0$. It also shows that the real distribution area corresponding to $c_1$ and $c_2$: $\rho_{S_1}$ and $\rho_{S_2}$ is almost equal in proportion and therefore equal to 1/2. 
    }
    \label{fig:theoretical}
\end{figure*}

\subsection{Theoretical Analysis}
The classifier $Q$ is trained only using Objective~\ref{eq:lc}, and is applied on the generated samples $\hat{x}$ as well as the real samples $x$ to decide the discriminator to use. It is optimal when it is able to correctly classify the generated samples $\hat{x}$ into their corresponding $c_i$'s. Empirically we observe that the classifier $Q$ is easily able to reach its optimum, as can be observed in the Figure~\ref{fig:theoretical}(b) and~\ref{fig:theoretical}(c), as the blue and orange curves (samples predicted as $c_1$ and $c_2$ respectively) coincide with the samples forming blue and green area (samples generated using $c_1$ and $c_2$ respectively). Interestingly, we observe that the classifier $Q$ is able to indirectly control the generator $G$ through the discriminators as $G$ groups its generations according to the code vectors $c_i$.

Here we provide formal theoretical formulation of our model with proof presented in Appendix~\ref{app:theo}. 

\begin{lemma}
For optimal $Q$ and fixed $G$, the optimal $D_i$, $\forall i \in [1,\dots, N]$ is
\begin{equation}
    D^{*}_i(x)=\frac{\rho_{S_i} p_d^i(x)}{\rho_{S_i} p_d^i(x)+\frac{1}{N}p_g^i(x)}
\end{equation}
where $S_i=\{x \in Supp(p_d)|Q(x)=c_i\}$, $\rho_{S_i} = \int_{x \in S_i} p_d(x) \mathrm{d}x$,  $p_d^i$ is a probability distribution such that $p_d^i(x) = \frac{p_d(x)}{\rho_{S_i} }$ and $Supp(p_d^i)=S_i$, and $p_g^i(x) = p_z(z)$ such that $G(z, c_i) = x$.
\end{lemma}
We can now reformulate the minimax game as $$U(G) = \mathop{\max}_{\{D_i\}_{i=1}^{N}}M(\{D_i\}_{i=1}^{N},G)$$
\begin{theorem}
In case of $N$ discriminators, the global minimum of $U(G)$ is achieved if and only if $p_g^i(x)=p_d^i(x)$, $\forall i \in [1,\dots,N]$. When $\rho_{S_i}=1/N$, the global minimum value of $U(G)$ is $-\log(4)$.
\end{theorem}

Sampling from $p_d^i$ is same as sampling from the $i^\text{th}$ mode of the real distribution, the mode that covers the set of samples $S_i=\{x \in Supp(p_d)|Q(x)=c_i\}$. Please note that we can assume that each of $\{p_d^i\}_{i=1}^N$ has a disjoint support. Figure~\ref{fig:theoretical}(b) and~\ref{fig:theoretical}(c) empirically show that the assumption of disjoint support of the distributions $p_d^1$ and $p_d^2$, which is decided by the classifier $Q$, is valid. 

So, in theory each $G_i(.)=G(.,c_i)$ should converge to a different mode as the target dataset distribution $p_d^i$ is itself different $\forall i \in [1,\dots, N]$. Hence, empirically the number of modes covered should essentially be at least more diverse than the standard GAN model. This is also observed in all our experiments as well as when comparing the Figures ~\ref{fig:theoretical}(a) and~\ref{fig:theoretical}(b).

\begin{corollary}
At global minimum of $U(G)$, the generative model $G$ replicates the real distribution $p_d$, categorized into different modes.
\end{corollary}

Thus our model \DPN{} can learn the real data distribution while also controlling the diversity of the generations by sampling from a different real mode corresponding to each $c_i$, which we also verify experimentally in the next section.

\input{experiments.tex}

\section{Discussion}
We conclude that it is not necessary for a generator to have equal capacity adversary to converge, meaning that the standard GAN training procedure could be enhanced with multiple (and even weaker) discriminators specialized only in attracting the model distribution of the generator to their corresponding modes.

\DPN{} is proven experimentally to utilize the capability of multiple discriminators by partitioning the target distributions into several identifiable modes and making each discriminator work on a separate mode. Thus, it reduces the complexity of the modes to be learnt by each discriminator. We show qualitatively and quantitatively that \DPN{} is able to better cover the real distribution. We observe that the generator is also able to sample from different identifiable modes of the data distribution given the corresponding code vectors.

\section*{Acknowledgement}
This work was supported by the ERC grant ERC-2012-AdG 321162-HELIOS, EPSRC grant Seebibyte EP/M013774/1 and EPSRC/MURI grant EP/N019474/1. We would also like to acknowledge the Royal Academy of Engineering and FiveAI. Viveka is wholly funded by Toyota Research Institute's grant.
\bibliography{gan}
\bibliographystyle{icml2019}

\input{appendix.tex}

\end{document}

%% file: experiments.tex
\section{Experiments}
\label{sec:exp}

We demonstrate the performance of our method \DPN{} on a diverse set of tasks with increasing complexity, involving probability density estimation and image generation. To illustrate the functioning of \DPN{}, we first set up two low-dimensional experiments (Section \ref{subsec:GMM}) using Gaussian Mixture Models (GMMs) as the target probability density function: 1D GMM and 2D GMM.
For the 1D Gaussian Mixture case, we compare \DPN{}'s robustness against other approaches by reproducing the experiment setting detailed in \cite{ghosh2017multi} and we outperform all competing methods both qualitatively and quantitatively. We also show \DPN{}'s performance using multiple discriminators and show how the training dynamics change according to the number of discriminators. We observe that increasing the number of discriminators improves the performance of the network until the point where the number of discriminators exceeds the number of underlying modes. Using the 2D circular GMM, we show that classifier $Q$ is able to learn good partitioning of the distribution and therefore each discriminator acts on samples from a different mode unlike GMAN~\cite{durugkar2016generative}. We show that \DPN{} is able to utilize the capacity of multiple discriminators and we can control the mode the generator samples from using the code $c$. Even in this case, \DPN{} performs better in capturing all the modes.

We finally demonstrate qualitative results on commonly investigated datasets: Stacked-MNIST, CIFAR-10 and CelebA in Section~\ref{subsec:img}. \DPN{} is able to generate good quality diverse samples. In case of CIFAR-10, we also show that we can generate samples from every class given the class label $y$. The information about the network architectures and the implementation details are provided in Appendix~\ref{app:imp_det}.

\subsection{Synthetic low dimensional distributions}
\label{subsec:GMM}
In \DPN{}, the role of the classifier $Q$ is to partition both the real and generated data-points into different clusters or modes, and each discriminator is consequently only trained on a separate cluster.
In order to fully understand how this helps the training, we experimented with two toy datasets obtained using mixture of Gaussian variants: a 1D GMM with 5 modes, as used in \cite{ghosh2017multi}, and a 2D circular GMM with 3 and 8 modes on the unit circle.

\begin{figure*}[!ht]
    \centering
    \includegraphics[width=\textwidth]{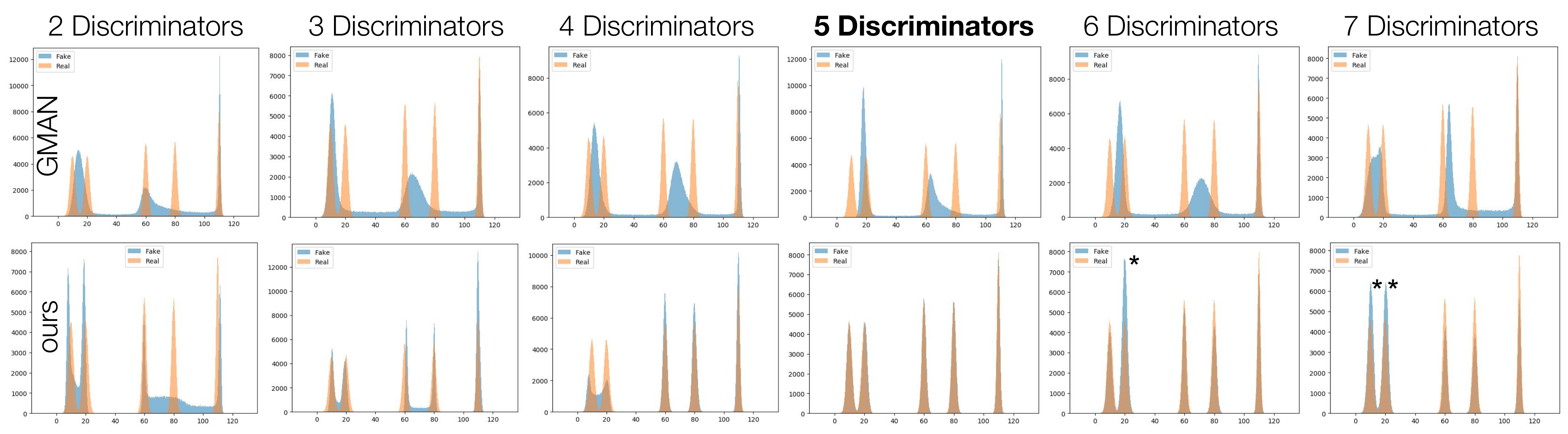}
    \caption{
        To study the behaviour of multi-discriminator settings with different number ($N$) of discriminators, we trained GMAN~\cite{durugkar2016generative} and \DPN{} on a 1D data set with 5 modes. Both GMAN's (top) and \DPN{}'s (bottom) results improved by adding discriminators while $N$ is less or equal the number of modes. To justify this, we point at the case where $N=5$: the models perform best since each class has to capture only a single mode. 
        In the case of $N=6$ and $N=7$ \DPN{} decreased in performance only due to oversampling (marked with $*$ and $**$) of some classes. 
    }
    \label{fig:1d}
\end{figure*}

    \subsubsection{Experiment setup and Evaluation details}
    First, we reproduced the 1D setting in \cite{ghosh2017multi} with 5 modes at $[10, 20, 60, 80, 110]$ and standard deviations $[3, 3, 2, 2, 1]$ respectively and we compare to the numbers reported in that paper in Table~\ref{tab:1d-var}. We sampled $65,536$ data points each from the real distribution and the generator distribution. For each of these two distributions, we created a histogram using bin size of $0.1$ with bins lying in the range of $-10$ to $130$. We then obtained Chi-square distance as well as KL divergence between the generator distribution and the true data distribution using these two histograms. To compare against GMAN using different number of discriminators, we used $1,000,000$ samples (instead of $65,536$ above) and show the results in Table~\ref{tab:1d-multi-d} and Figure~\ref{fig:1d}.

    We then introduce a 2D experiment setting with 2D Gaussian Mixture Model (GMM). It has multiple modes having covariance matrix of $0.01I$, where $I$ is an identity matrix, and equally separated means lying on a unit circle (please refer to Figure~\ref{fig:2d} for the $3$ mode case). For Table~\ref{tab:2d} we consider $8$ modes and construct histograms using $1,000,000$ samples and bin size of $0.0028\times 0.0028$ with bins lying in the range of $[-1.4, 1.4]\times [-1.4,1.4]$.

    For these experiments, we use uniform distribution $U(-1, 1)$ of dimension $64$ for $p_z$ and uniform categorical distribution for $p_c$ to get the generations in both 1D and 2D experiments.
    \input{tables/1d-var.tex}
    \subsubsection{Observations}
    \label{sub:obs}
        \paragraph{Comparing against other GAN variants}
        In Table \ref{tab:1d-var}, we show that \DPN{} outperforms other GAN architectures on the 1D task by a large margin in terms of Chi-square distance and KL-Divergence.
        We believe that the success is due to the classifier $Q$'s capability to learn to partition the underlying distribution easily. 
        We also show in Table \ref{tab:2d}, that in the 2D task \DPN{} achieves better performance than GMAN~\cite{durugkar2016generative} in terms of both  KL-Divergence an Chi-square.
        
        \paragraph{Benchmarking the number of discriminators}
        \begin{figure*}[!htb]
            \centering
            \includegraphics[width=0.95\textwidth]{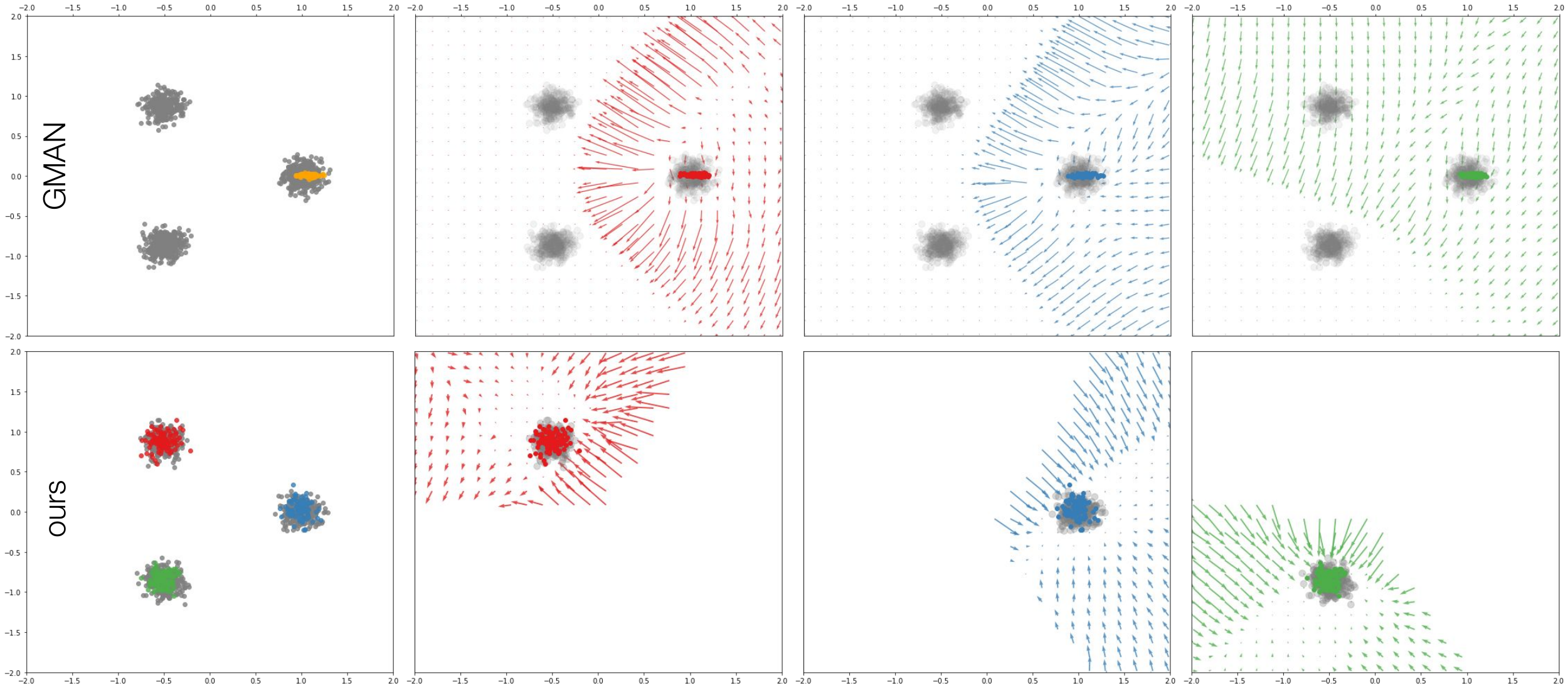}
            \caption{
                A circular 2D GMM with 3 modes on the unit circle with standard deviations of 0.01. We train GMAN and \DPN{} with $N=3$ setting. 
                In the first column we show the generated distribution for each method in colours (orange for GMAN, a single color since GMAN uses no conditioning on their samples, and red, green and blue in \DPN{}'s case for samples generated using $c_1$, $c_2$ and $c_3$ respectively).
                In columns 2, 3 and 4 we plotted the gradient field for each of the equally separated data-points present in $[-2.0, 2.0]\times [-2.0, 2.0]$. Different colors show the gradient by different discriminators. For GMAN, each column shows the gradient field by the respective discriminator for each of the data-points. For \DPN{}, each of the data-points is first classified by $Q$ and the respective discriminator is used to get the gradient field, which is shown in the corresponding column of 2-3.
                We see how classifier $Q$ indirectly pushes different modes generated by $G$ apart, more importantly the gradient field has the highest magnitude in the direction which separates the modes, while this phenomenon is not happening in the case of GMAN.
                \vspace{-10pt}
            }
            \label{fig:2d}
        \end{figure*}
        \input{tables/1d-multi-d.tex}
        We study the change in performance with regards to the number of discriminators used by both GMAN \cite{durugkar2016generative} and \DPN{}. 
        The clustering mechanism with varying number of discriminators is illustrated on the 1D task in Figure \ref{fig:1d}. 
        We see in this experiment that classes of the generated samples are first attracted towards larger clusters of the real data. By adding more discriminators, the quality of the reconstructed modes is refined.
        The refinement process starts first with the easiest separation, between the $2^{nd}$ and the $3^{rd}$ peaks, after that the $4^{th}$ and $5^{th}$ modes are distinguished by the classifier $Q$, and so on.
        We quantitatively see in Table \ref{tab:1d-multi-d}  that increasing the number of discriminators improves the performance of both GMAN and \DPN{} up to a certain point where $N$ (number of discriminators) matches the number of modes in the data. 
        After this optimal point, increasing $N$ yields a decreasing performance, because already captured modes are oversampled.
        In Figure \ref{fig:1d} we have marked examples of oversampling in the last two columns with $*$ symbols.
        \input{tables/2d.tex}
        It is interesting to note that when the same experiment was carried out in MADGAN~\cite{ghosh2017multi}, which uses multiple generators, 
        their performance peaked at $N_{Generators}=4$ unlike GMAN and ours, both of which logically peaked at $N_{Discriminator}=5$ considering that there are $5$ visible modes. This shows a difficulty in tuning the hyper-parameter $N_{Generators}$ in \cite{ghosh2017multi} for different applications.
        \paragraph{2D experiments}
        In 2D experiments, for both GMAN~\cite{durugkar2016generative} and \DPN{} we experiment with $N=M=8$ (where $N$ is the number of discriminators, $M$ is the number of modes) for both quantitative (listed in Table \ref{tab:2d}) and qualitative results (see Appendix~\ref{app:2d}), and the $N=M=3$ setting for qualitative results (illustrated in Figure \ref{fig:2d}).
        In all of the runs, \DPN{} was able to capture, and classify all modes of the true distribution correctly, while GMAN~\cite{durugkar2016generative} failed on both the $N=M=3$ as well as the $N=M=8$ setting.
        
        In Figure ~\ref{fig:2d} we show a circular 2D GMM with 3 modes on the unit circle which is used to train GMAN and DoPaNets.
        In the case of \DPN{}, it can also be observed (see column 1) that the generator generates from a different mode for a different $c_i$. We can also visually see that the classifier $Q$ is indirectly able to control the conditioned samples $G(z, c)$ by routing them to the corresponding discriminators (see columns 2-3). It also illustrates that we are indeed able to utilize the capabilities of multiple $D_i$'s as intended: different discriminators begin to specialize on different modes and therefore provide different gradients for the respective mode as well. Although being trained with the generated code vectors only, \DPN{}'s classifier $Q$ achieves fine partitioning of the original distribution. 
        We suggest that our approach succeeds because each discriminator is fed different samples from the beginning.
        $Q$ is initialized to assign each real sample to every discriminator with equal probability, but given that the generator samples different points for every code vector $Q$ quickly learns the different modes that the samples from $G_i$ are attracted towards (where $G_i$ refers to the conditional distribution modeled by $G(z,c_i)$).
        Given that the updated $Q$ is already providing different subsets of the input space to the different discriminators, the discriminators will provide different gradients for each corresponding code vector.
        Therefore $G$ learns to separate the modes of the learnt distributions conditioned on $c$ from each other. We argue that GMAN is not able to utilize multiple discriminators in this experiment setup and that most of the learning is done by just a few discriminators rather than their effective ensemble (see Appendix~\ref{app:2d}).

\subsection{Image generation}
\label{subsec:img}
After investigating the \DPN{} performance on low dimensional tasks, now we validate \DPN{} on real image generation tasks.

    \subsubsection{Stacked-MNIST}
\begin{figure}[!ht]
    \begin{minipage}{\linewidth}
    \centering{
        \subfloat[]{\includegraphics[width=.45\linewidth]{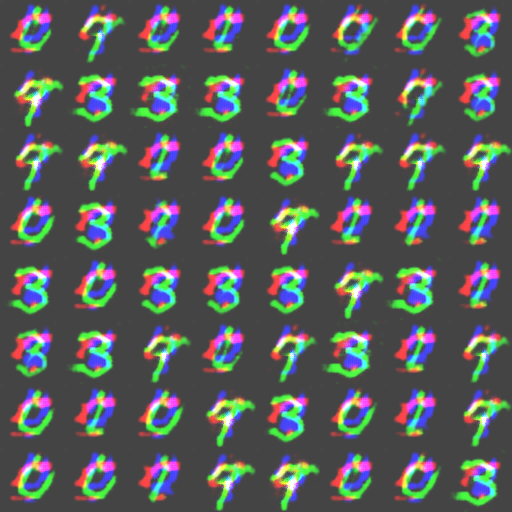}}~
        \subfloat[]{\includegraphics[width=.45\linewidth]{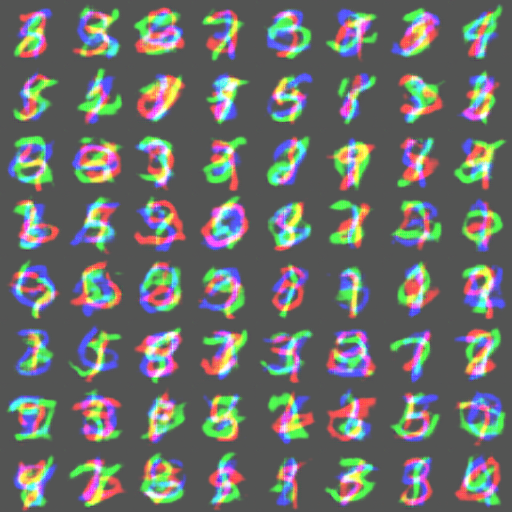}}
        \caption{
            Sample images from stacked mnist. The samples on the left (a) were generated using the traditional architecture of DCGAN\footnotemark while the samples on the right (b) were generated using \DPN{} with $10$ discriminators ($N=10$). 
            Visually, (a) appears to be more clear as only one color (green) is dominating over digits of the other two colors while in (b) more digits are covered per sample which refers to higher diversity in the modeled distribution, thus better mode coverage.
        }\label{FigStacedMnistSamples}
    }
    \end{minipage}
    \end{figure}

    We first investigate how well \DPN{} can reconstruct the real distribution of the data using the Stacked-MNIST dataset \cite{srivastava2017veegan}. 
    This dataset contains three channel color images, containing a randomly selected sample from the MNIST dataset in each channel. This results in ten possible modes on each channel so the number of all the possible modes in the dataset is $10^3$. 
    It was shown in \cite{ghosh2017multi} that various architectures recovered only a small portion of these modes. 
    A qualitative image depicting the recovered modes using the traditional DCGAN~\cite{Radford16} architecture and \DPN{} can be seen in Figure \ref{FigStacedMnistSamples}.
    We have also measured the Kullback-Leibler divergence between the real distribution and the generated distributions. We compare \DPN{} against the other GAN variants in Table~\ref{tab:stacked_mnist}.
    \input{tables/stacked-mnist.tex}
    
    \footnotetext{Implementation details can be found at: \href{https://github.com/carpedm20/DCGAN-tensorflow}{https://github.com/carpedm20/DCGAN-tensorflow}}
    \subsubsection{Qualitative results on CIFAR-10 and celebA}
    \label{subsubsec:qualImg}
\begin{figure}
    \centering
    \includegraphics[width=.45\textwidth]{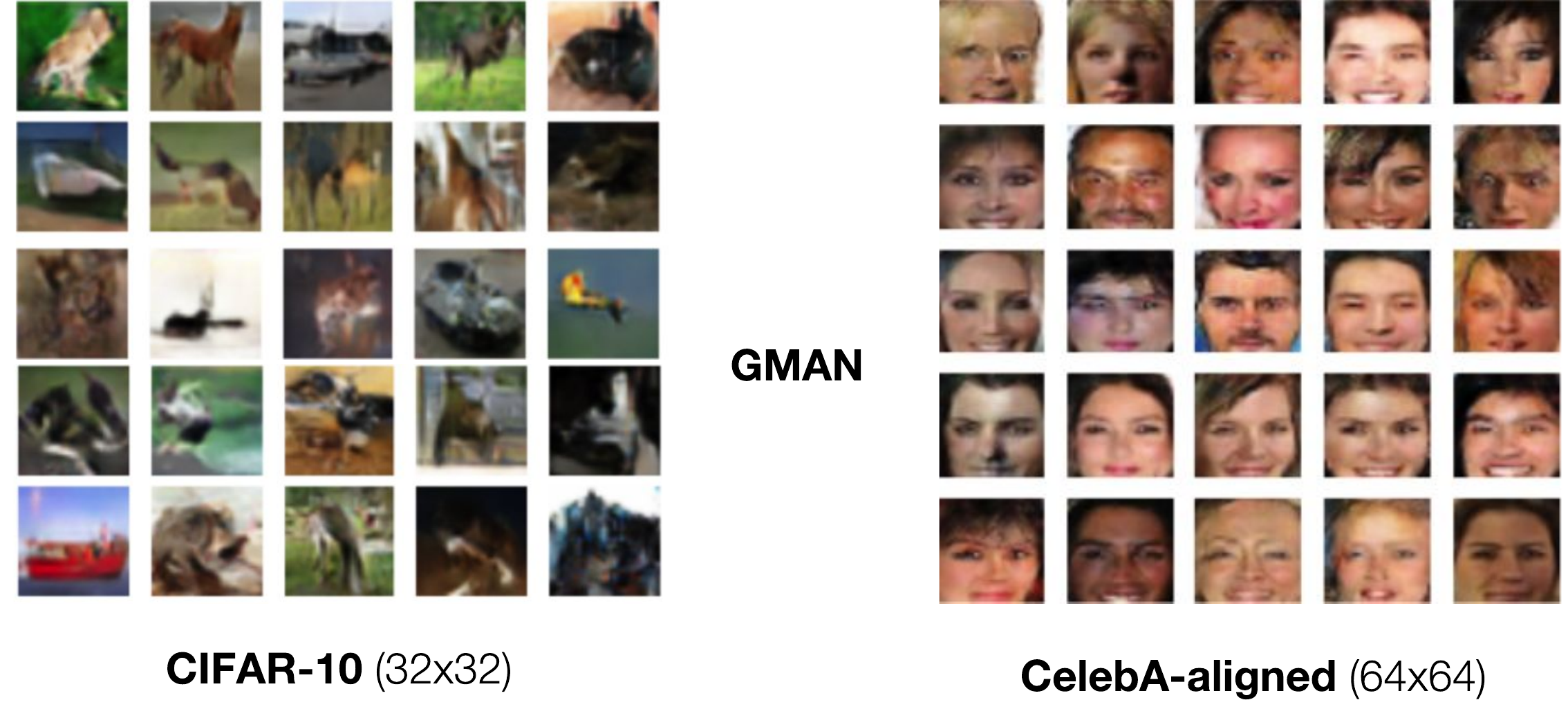}
    \caption{Random samples presented in \cite{durugkar2016generative} for image generation tasks such as CIFAR-10 and CelebA. 
    Best results were achieved with GMAN-0 variant with $N=5$ discriminators. 
    For CelebA they cropped the images to exclude background.
    }
    \label{fig:GMAN-compare}
    \vspace{-10pt}
\end{figure}

    To show the image generation capabilities of \DPN{}, we trained the multi discriminator setting on a lower and a higher complexity image generation task, CIFAR-10 and CelebA respectively. We compare our results qualitatively to the ones reported by GMAN~\cite{durugkar2016generative} on both tasks in Figure~\ref{fig:GMAN-compare}.

        \paragraph{CIFAR-10}
        \begin{figure}[ht!]
            \centering
            \includegraphics[width=0.3\textwidth]{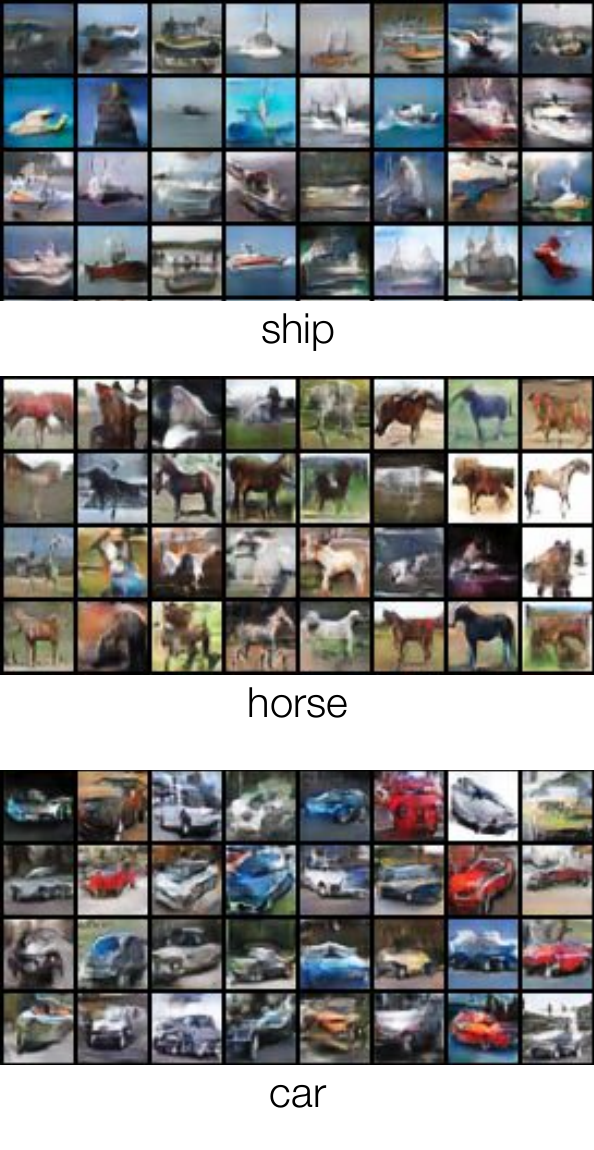}
            \caption{
            $32\times 32$ CIFAR-10 samples drawn from \DPN{} trained with $N=5$ discriminators for 700k iterations. The generations in each subfigure correspond to $y=1,2$ and $3$ respectively.
            We call the attention to the various different details learnt for each class which can cause mode collapse in other GAN variants.
            }
            \label{fig:cifar}
        \end{figure}
        While learning the distribution of $32\times 32$ colored images may sound easy, the main challenge is to learn geometric structures from low resolution and reproduce them in various colors, backgrounds, angles etc. Following~\cite{Mescheder2018ICML}, the generator takes ground truth label $y$ as input as well along with the code $c$, and the discriminator outputs a $10$ dimensional output of which only the $y^\text{th}$ index is used for training $D_\sigma$ as well as $G$, while $Q$ is trained just using the code $c$. Thus, the code $c$ helps it learn class invariant features.
        We illustrate in Figure \ref{fig:cifar} that \DPN{} is capable of capturing these features such as different object orientations and colors depicted in various weather conditions. It is also able to recognize minute details like wheels, horse hair, ship textures, etc. 
        We present more generations corresponding to each of the classes in Appendix~\ref{app:cifar}.
    
\begin{figure}[ht!]
    \centering
    \includegraphics[width=0.4\textwidth]{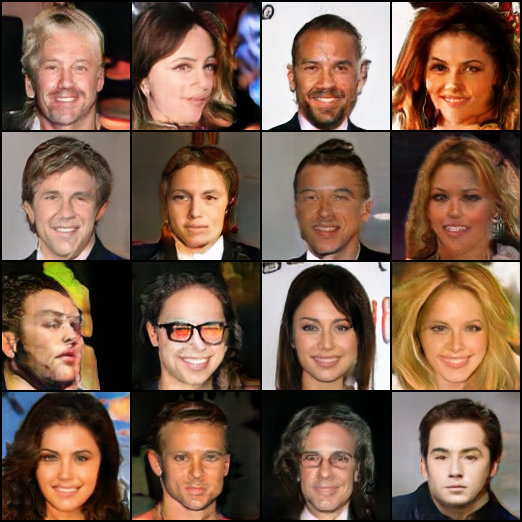}
    \caption{Random samples from \DPN{} trained at scale of $128\times 128$ images on unaligned CelebA set for 100k iterations.
    As it can be seen, different faces appear in diverse poses with different background, and rarely occurring accessories such as orange sunglasses are learned by the model.}
    \label{fig:celeb}
    \vspace{-5pt}
\end{figure}

        \paragraph{CelebA}
        We also show \DPN{} performance on large scale images such as $128\times 128$ by training a residual network for 100k iterations on the celebA dataset. 
        This dataset contains various modes like lighting, pose, gender, hair style, clothing, facial expressions which are challenging to capture for generative models.
        In Figure \ref{fig:celeb} we demonstrate that \DPN{} is capable of recovering the aforementioned visual features.
    
%% file: tables/1d-var.tex
\begin{table}
\centering
\begin{tabular}{@{}ccc@{}}
\toprule
\textbf{GAN Variants} & \textbf{Chi-square}($\times10^5$) & \textbf{KL-Div} \\ \midrule
DCGAN$^*$                 & 0.90                                & 0.322           \\
WGAN$^*$                  & 1.32                                & 0.614           \\
BEGAN$^*$                 & 1.06                                & 0.944           \\
GoGAN$^*$                 & 2.52                                & 0.652           \\
Unrolled GAN$^*$          & 3.98                                & 1.321           \\
Mode-Reg DCGAN$^*$        & 1.02                                & 0.927           \\
InfoGAN$^*$               & 0.83                                & 0.210           \\
MA-GAN$^*$                & 1.39                                & 0.526           \\
MAD-GAN$^*$               & 0.24                                & 0.145           \\
GMAN                  & 1.44                                & 0.69            \\
\textbf{\DPN{}}         & \textbf{0.03}                       & \textbf{0.02}   \\ \bottomrule
\end{tabular}
\caption{1D Gaussian Mixture Model experiment using best results from 3 runs for GAN variants that aims to solve mode collapse. Results for the GAN variants marked as $*$ were reproduced from~\cite{ghosh2017multi}.\vspace{-10pt}}
\label{tab:1d-var}
\end{table}

%% file: tables/1d-multi-d.tex
\begin{table}[!ht]
\centering
\begin{tabular}{@{}ccccc@{}}
\toprule
\textbf{$N$} & \multicolumn{2}{c}{\textbf{Chi-square}($\times10^7$)} & \multicolumn{2}{c}{\textbf{KL-Div}} \\ \midrule
                          & GMAN                       & \textbf{\DPN{}}                       & GMAN             & \textbf{\DPN{}}      \\
2 & 5.00$\pm$6.80 & \textbf{1.89$\pm$0.92} & 1.74$\pm$0.63 & \textbf{0.81$\pm$0.27} \\
3 & 2.96$\pm$2.88 & \textbf{1.10$\pm$2.43} & 1.50$\pm$0.57 & \textbf{0.55$\pm$0.36} \\
4 & 3.41$\pm$2.73 & \textbf{0.74$\pm$0.98} & 1.48$\pm$0.42 & \textbf{0.50$\pm$0.41} \\
5 & 4.62$\pm$3.92 & \textbf{0.27$\pm$0.54} & 1.55$\pm$0.30 & \textbf{0.25$\pm$0.26} \\
6 & 3.94$\pm$3.22 & \textbf{0.41$\pm$0.50} & 1.56$\pm$0.22 & \textbf{0.35$\pm$0.20} \\
7 & 2.84$\pm$1.51 & \textbf{0.42$\pm$0.43} & 1.45$\pm$0.38 & \textbf{0.36$\pm$0.21} \\
8 & 2.80$\pm$1.55 & \textbf{0.93$\pm$1.14} & 1.36$\pm$0.43 & \textbf{0.56$\pm$0.31} \\

\bottomrule

\end{tabular}
\caption{1D Gaussian Mixture Model experiment using best results from 20 runs with different number of discriminators ($N$) as illustrated in Figure~\ref{fig:1d}.}
\label{tab:1d-multi-d}
\end{table}

%% file: tables/2d.tex
\begin{table}
\centering
\begin{tabular}{@{}ccc@{}}
\toprule
\textbf{GAN Variants} & \textbf{Chi-square}($\times10^6$) & \textbf{KL-Div} \\ \midrule
Standard GAN          & 3.883                               & 2.860           \\
GMAN                  & 1.253                               & 0.636           \\
\textbf{\DPN{}}         & \textbf{0.449}                      & \textbf{0.246}  \\ \bottomrule
\end{tabular}
\caption{
        2D Gaussian Mixture Model experiment with $M=N=8$ (where $N$ is the number of discriminators, $M$ is the number of Gaussians used in the mixture model).
        For each experiment, we use a fixed set of 1,000,000 samples and do 5 runs for each algorithm and report the results using the best run.
        We took effort to make sure that the comparison was fair, and used the same set of parameters as in the 1D experiments.\vspace{-10pt}}
\label{tab:2d}
\end{table}

%% file: tables/stacked-mnist.tex
\begin{table}
	\centering
    \begin{tabular}{@{}cc@{}} 
		\toprule
		\textbf{GAN Variants} & \textbf{KL Div} \\
		\midrule
		DCGAN$^*$~\cite{Radford16}  & 2.15\\ 
		WGAN$^*$~\cite{arjovsky2017wasserstein}  & 1.02\\ 
		BEGAN$^*$~\cite{berthelot2017began}  & 1.89\\ 
		GoGAN$^*$~\cite{juefei2017gang}  & 2.89\\
		Unrolled GAN$^*$~\cite{metz2016unrolled} & 1.29\\ 
		Mode-Reg DCGAN$^*$~\cite{che2016mode}  & 1.79\\ 
		InfoGAN$^*$~\cite{chen2016infogan}  & 2.75\\ 
		MA-GAN$^*$~\cite{ghosh2017multi} & $3.4$\\
		MAD-GAN$^*$~\cite{ghosh2017multi} &  0.91\\ 
		GMAN~\cite{durugkar2016generative} & 2.17\\
		\textbf{\DPN{} (ours)} & \textbf{0.13}\\
		\bottomrule		
	\end{tabular}
	\caption{Stacked-MNIST: we compare our method against several GAN variants. Through this experiment using a real dataset, we can show that \DPN{} is closer to the real distribution. Results for the GAN variants marked as $*$ were reproduced from~\cite{ghosh2017multi}.}
	\label{tab:stacked_mnist}
\end{table}

%% file: appendix.tex
\clearpage
\appendix

\section*{Appendix}
Here we first give the theoretical formulation of our work \DPN{} to show that the modes captured should be different for each categorical code $c_i$ and that the standard GAN can be considered as its lower bound on mode collapse. We then give some more experimental insights from the 2D task. We also give some more generations using CIFAR. Later we provide implementation details of the network architectures we used.

\section{Theoretical formulation}
\label{app:theo}
Here we present the theoretical formulation for our proposed method \DPN{}.

\begin{lemma1}
For optimal $Q$ and fixed $G$, the optimal $D_i$, $\forall i \in [1,\dots, N]$ is
\begin{equation}
    D^{*}_i(x)=\frac{\rho_{S_i} p_d^i(x)}{\rho_{S_i} p_d^i(x)+\frac{1}{N}p_g^i(x)}
\end{equation}
where $S_i=\{x \in Supp(p_d)|Q(x)=c_i\}$, $\rho_{S_i} = \int_{x \in S_i} p_d(x) \mathrm{d}x$,  $p_d^i$ is a probability distribution such that $p_d^i(x) = \frac{p_d(x)}{\rho_{S_i} }$ and $Supp(p_d^i)=S_i$, and $p_g^i(x) = p_z(z)$ such that $G(z, c_i) = x$.
\end{lemma1}

\begin{proof}
Let us consider a case where we have $N=2$ discriminators. The theoretical formulation for this case can be trivially extended to more number of discriminators. The objective being optimized by the generator and the discriminators is (Obj.~\ref{eq:minmax}):

\begin{equation}\begin{split}\label{eq:ls_c_prel}
        \mathop{\min}_{G}\mathop{\max}_{D_1, D_2}&M(\{D_1, D_2\},G) := \Bigg[\mathop{\mathbb{E}}_{\substack{x \sim p_{d} \\ \sigma \sim Q(x)}}\left[\log D_\sigma \left(x\right)\right]\\
        +& \mathop{\mathbb{E}}_{\substack{z\sim p_{z}, c\sim p_{c} \\ \hat\sigma \sim Q(G(z, c)) }}
        \left[\log \left(1-D_{\hat{\sigma}}\left(G\left({z}, {c}\right)\right)\right)\right]
        \Bigg]
\end{split}
\end{equation}

When the classifier $Q$ has converged to its optimal form, the above Equation \ref{eq:ls_c_prel} can be rewritten as:

\begin{equation}\begin{split}\label{eq:ls_c}
        \mathop{\min}_{G}\Bigg[\mathop{\max}_{D_1} \Big[&\mathop{\mathbb{E}}_{\substack{x \sim p_{d} \\ x \in S_1}}\left[\log D_1 \left(x\right)\right]\\
        &+ p_c(c_1)\mathop{\mathbb{E}}_{z \sim p_{z}}\left[\log \left(1-D_1\left(G\left({z,c_1}\right)\right)\right)\right]\Big]\\
        + \mathop{\max}_{D_2} \Big[&\mathop{\mathbb{E}}_{\substack{x \sim p_{d} \\ x \in S_2}}\left[\log D_2 \left(x\right)\right]\\
        &+ p_c(c_2)\mathop{\mathbb{E}}_{z \sim p_{z}}\left[\log \left(1-D_2\left(G\left({z,c_2}\right)\right)\right)\right]
        \Big]\Bigg]
\end{split}
\end{equation}
where $x \in S_i$ if $Q(x)=c_i$. $p_c$ is the categorical distribution and in our case equal probability is assigned to both the values $c_1$ and $c_2$. Here $c_i$ is that code vector which leads the classifier $Q$ to pass $G\left({z,c_i}\right)$ to $D_i$ for $i \in [1,2]$. Please note that we can therefore consider $G\left({.,c_1}\right)$ as $G_1(.)$ and $G\left({.,c_2}\right)$ as $G_2(.)$, where $G_1$ and $G_2$ have shared weights except the bias weights in the initial layer. Bias weights in the initial layer are independently trained for $G_1$ and $G_2$.

The Objective \ref{eq:ls_c} can be rewritten as:
\begin{equation}\begin{split}\label{eq:ls_c_splitG}
        \mathop{\min}_{G_1, G_2}\Bigg[\mathop{\max}_{D_1}\Big[&\rho_{S_1}\mathop{\mathbb{E}}_{x \sim p_d^1}\left[\log D_1 \left(x\right)\right]\\
        &+ \frac{1}{2}\mathop{\mathbb{E}}_{z \sim p_{z}}\left[\log \left(1-D_1\left(G_1\left({z}\right)\right)\right)\right]\Big]\\
        + \mathop{\max}_{D_2}\Big[&\rho_{S_2}\mathop{\mathbb{E}}_{x \sim p_d^2}\left[\log D_2 \left(x\right)\right]\\
        &+ \frac{1}{2}\mathop{\mathbb{E}}_{z \sim p_{z}}\left[\log \left(1-D_2\left(G_2\left({z}\right)\right)\right)\right]\Big]\Bigg]
\end{split}
\end{equation}

where $\rho_{S_i} = \int_{x \in S_i} p_d(x) \mathrm{d}x$. $p_d^i$ is a probability distribution such that $p_d^i(x) = \frac{p_d(x)}{\rho_{S_i} }$ and $Supp(p_d^i)=S_i$ where $S_i=\{x \in Supp(p_d)|Q(x)=c_i\}$ is the set of samples in the $i^\text{th}$ mode of the real distribution. So, sampling from $p_d^i$ is same as sampling from the $i^\text{th}$ mode of the real distribution $p_d$. Therefore, $Supp(p_d)=Supp(p_d^1) \cup Supp(p_d^2)$ and $Supp(p_d^1) \cap Supp(p_d^2) = \emptyset$.

For a fixed generator $G$, $G_1$ and $G_2$ are also fixed. For a given $G_1$ and $G_2$, the discriminator $D_i$ tries to maximize the quantity (using Objective~\ref{eq:ls_c_splitG}):
\begin{equation}\begin{split}
    \rho_{S_i}\int_x p_d^i(x)\log D_i(x) &\mathrm{d}x \\+ \frac{1}{2}\int_z&p_z(z)\log(1-D_i(G_i(z)))\mathrm{d}z\\
    =\int_x \rho_{S_i} p_d^i(x)\log D_i(x)& + \frac{1}{2}p_g^i(x)\log(1-D_i(x))\mathrm{d}x
\end{split}
\end{equation}

where $p_g^i(x) = p_z(z)$ such that $G_i(z) = x$ for $i = 1, 2$. Therefore, for a fixed generator we get the optimal discriminator $D_i$ as:
\begin{equation}\label{eq:opt-Di}
D^{*}_i(x)=\frac{\rho_{S_i} p_d^i(x)}{\rho_{S_i} p_d^i(x)+\frac{1}{2}p_g^i(x)}
\end{equation}

In case of $N$ discriminators, the optimal discriminator $D_i$ can be similarly obtained as:
\begin{equation}\label{eq:opt-Di-general}
D^{*}_i(x)=\frac{\rho_{S_i} p_d^i(x)}{\rho_{S_i} p_d^i(x)+\frac{1}{N}p_g^i(x)}
\end{equation}
\end{proof}

\begin{theorem2}
In case of $N$ discriminators, the global minimum of $U(G)$ is achieved if and only if $p_g^i(x)=p_d^i(x)$, $\forall i \in [1,\dots,N]$. When $\rho_{S_i}=1/N$, the global minimum value of $U(G)$ is $-\log(4)$.
\end{theorem2}

\begin{proof}
Given the optimal discriminators $D_1^*$ and $D_2^*$, we can reformulate the Objective \ref{eq:ls_c_splitG} as:

\begin{equation}\begin{split}\label{eq:ls_c1}
        \mathop{\min}_{G_1, G_2}\Bigg[\rho_{S_1}&\mathop{\mathbb{E}}_{x \sim p_d^1}\left[\log D^*_1 \left(x\right)\right]\\
        +& \frac{1}{2}\mathop{\mathbb{E}}_{z \sim p_{z}}\left[\log \left(1-D^*_1\left(G_1\left({z}\right)\right)\right)\right]\\
        + \rho_{S_2}&\mathop{\mathbb{E}}_{x \sim p_d^2}\left[\log D^*_2 \left(x\right)\right]\\
        +& \frac{1}{2}\mathop{\mathbb{E}}_{z \sim p_{z}}\left[\log \left(1-D^*_2\left(G_2\left({z}\right)\right)\right)\right]\Bigg]
\end{split}
\end{equation}

As noted earlier, bias weights in the initial layer of $G_1$ and $G_2$ are independently trained with all the other weights shared. As it empirically turns out, the shared weights help learn the similar features, which are essential in low-level image formation and should be similar even if $G_1$ and $G_2$ were trained independently. So, we can rather relax the restriction and consider $G_1$ and $G_2$ to be independent of each other. So, the objective \ref{eq:ls_c1} can be rewritten as:

\begin{equation}
    \label{eq:ls_c2}
    \mathop{\min}_{G_1}W(G_1)+\mathop{\min}_{G_2}W(G_2)
\end{equation}

where,
\begin{equation}
 \label{eq:ls_c_splitW}
\begin{split}
    W(G_i):=\Bigg[\rho_{S_i}&\mathop{\mathbb{E}}_{x \sim p_d^i}\left[\log D^*_i \left(x\right)\right]\\
        +& \frac{1}{2}\mathop{\mathbb{E}}_{z \sim p_{z}}\left[\log \left(1-D^*_i\left(G_i\left({z}\right)\right)\right)\right]\Bigg]
        \end{split}
\end{equation}

This is same as optimizing different $G_i$-$D_i$ pairs on dataset distributions $p_d^i$ decided by the classifier $Q$ based on the target real distribution $p_d$. Figure~\ref{fig:theoretical}(b) and~\ref{fig:theoretical}(c) empirically show that the assumption of disjoint support of the distributions $p_d^1$ and $p_d^2$ is valid. The Equation~\ref{eq:ls_c_splitW} can be rewritten as:

\begin{equation}
\label{eq:ls_c_splitW_2}
\begin{split}
    W(G_i):=\Bigg[\rho_{S_i}&\mathop{\mathbb{E}}_{x \sim p_d^i}\left[\log D^*_i \left(x\right)\right]\\
        +& \frac{1}{2}\mathop{\mathbb{E}}_{x \sim p_g^i}\left[\log \left(1-D^*_i\left(x\right)\right)\right]\Bigg]\\
        =\Bigg[\rho_{S_i}&\mathop{\mathbb{E}}_{x \sim p_d^i}\left[\log \frac{\rho_{S_i} p_d^i(x)}{\rho_{S_i} p_d^i(x)+\frac{1}{N}p_g^i(x)}\right]\\
        +& \frac{1}{2}\mathop{\mathbb{E}}_{x \sim p_g^i}\left[\log \left(\frac{\frac{1}{N}p_g^i(x)}{\rho_{S_i} p_d^i(x)+\frac{1}{N}p_g^i(x)}\right)\right]\Bigg]
\end{split}
\end{equation}

We can further reformulate Equation~\ref{eq:ls_c_splitW_2} as:

\begin{equation}
    \label{eq:wg_i_KL}
    \begin{split}
     W(G_i):=\rho_{S_i}\Bigg[&-\log(c_1^i)+KL\left(p_d^i\left \|\frac{p_d^i}{c_1^i}+\frac{p_g^i}{2\rho_{S_i}c_1^i}\right. \right)\\
     &+c_2^i\Bigg(-\log(c_3^i)\\
     &+KL\left(\frac{p_g^i}{2\rho_{S_i}c_2^i}\left \|\frac{p_d^i}{c_2^ic_3^i}+\frac{p_g^i}{2\rho_{S_i}c_2^ic_3^i}\right. \right)\Bigg)\Bigg]
    \end{split}
\end{equation}

where $KL$ is the Kullback-Leibler divergence, $c_1^i$, $c_2^i$ and $c_3^i$ are constants such that $2\rho_{S_i}c_2^i = 1$ ($\implies c_2^i=1/2\rho_{S_i}$) for $p_g^i/2\rho_{S_i}c_2^i$ (the first distribution of second $KL$ term) to be a probability distribution. The Kullback-Leibler divergence between two distributions is always non-negative and, zero iff the two distributions are equal. In above equation, the two $KL$ terms are zero simultaneously when $(c_1^i-1)(c_2^i-1)=1$ and the generator distribution is $$p_g^i=2(c_1^i-1)\rho_{S_i}p_d^i$$ where $2(c_1^i-1)\rho_{S_i} = 1$ ($\implies c_1^i=1+1/2\rho_{S_i}$) for $p_g^i$ to be a probability distribution. Therefore, the global minimum of Eq.~\ref{eq:wg_i_KL} is achieved iff $p_g^i=p_d^i$. The constants in Eq.~\ref{eq:wg_i_KL} are chosen such that:
\begin{equation}
    c_1^i=1+\frac{1}{2\rho_{S_i}},\qquad c_2^i=\frac{1}{2\rho_{S_i}},\qquad c_3^i=1+2\rho_{S_i}
\end{equation}

Please note that when $\rho_{S_i}=1/2$, the Eq.~\ref{eq:wg_i_KL} can be reformulated as:
\begin{equation}
     W(G_i):=-\log(2)+JSD\left(p_\text{data}\left \|p_g\right. \right)
\end{equation}

and the global minimum of $U(G)$ obtained is $-\log(4)$. This global minimum value is the same in general case for $N$ discriminators when $\rho_{S_i}=1/N$.
\end{proof}

\begin{corollary3}
At global minimum of $U(G)$, the generative model $G$ replicates the real distribution $p_d$, categorized into different modes.
\end{corollary3}
\begin{proof}
As noted in the proof of Lemma 3.1, sampling from $p_d^i$ is same as sampling from the $i^\text{th}$ mode of the real distribution $p_d$. At global minimum of $U(G)$, we have $p_g^i=p_d^i$ so $G_i(.)=G(.,c_i)$ is able to sample from the $i^\text{th}$ mode of the real distribution. As the real distribution is categorized into $N$ modes in total and each of $\{G(.,c_i)\}_{i=1}^N$ can samples from the corresponding modes, so $G$ can replicate the real distribution $p_d$, categorized into different modes.
\end{proof}

\section{2D GMM}
\label{app:2d}
\begin{figure}[!ht]
    \centering{
    \subfloat[]{
        \includegraphics[width=.85\linewidth]{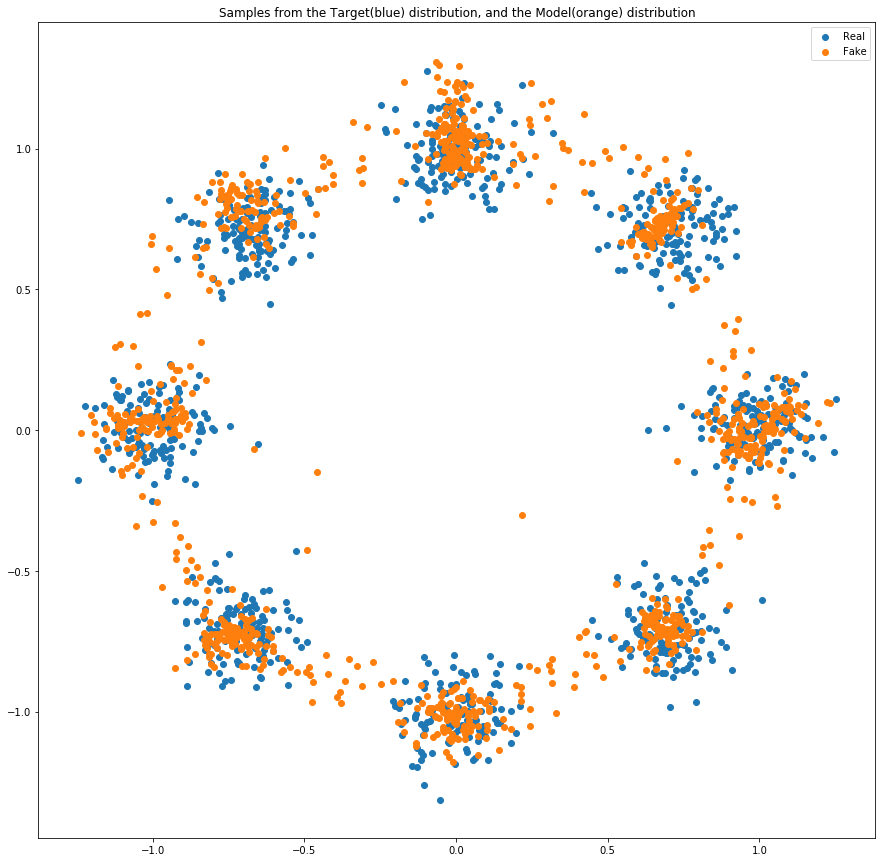}
    }\\
    \subfloat[]{
        \includegraphics[width=.85\linewidth]{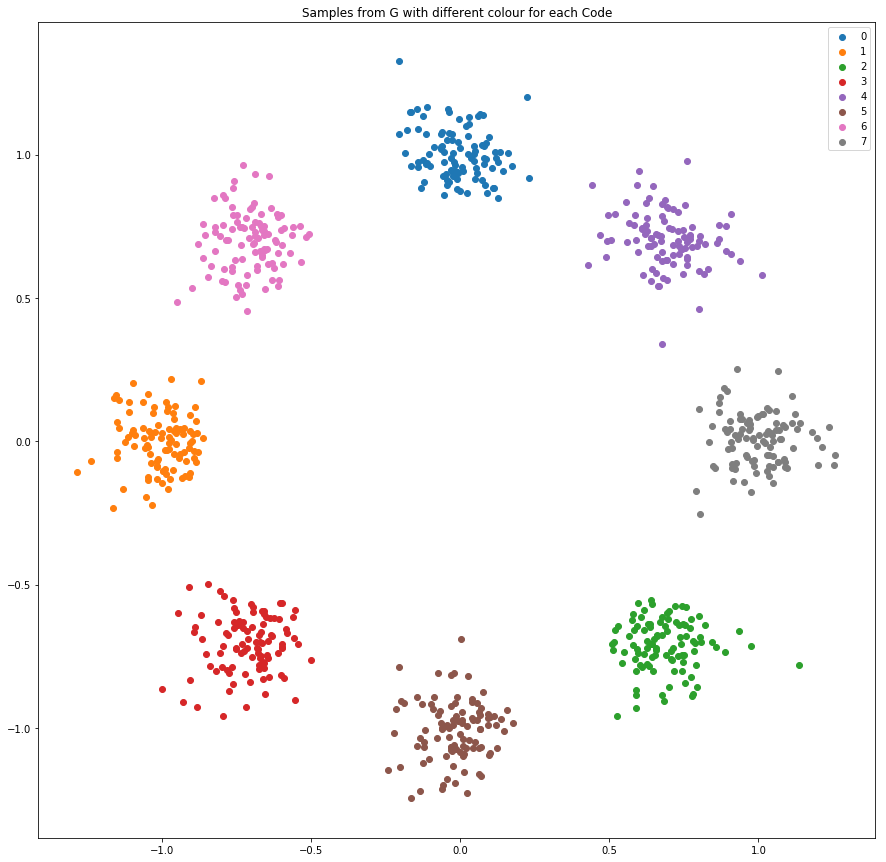}
    }
    
    \caption{
        (a) GMAN: scatter plot of the data distribution (blue) and model distribution (orange). 
        Although it is covering all the modes, there are a lot of false positives lying between the actual modes. 
        (b) \DPN{}: scatter plot of the model distribution with different colors assigned for samples generated using different code. We sample $c$ from uniform categorical distribution. Notice that the modes are clearly separated from each other as compared to GMAN.
    }
    \label{fig:8GMM}
    }
\end{figure}

\begin{figure*}[!ht]
    \centering{
    \subfloat[]{
        \includegraphics[width=.85\linewidth]{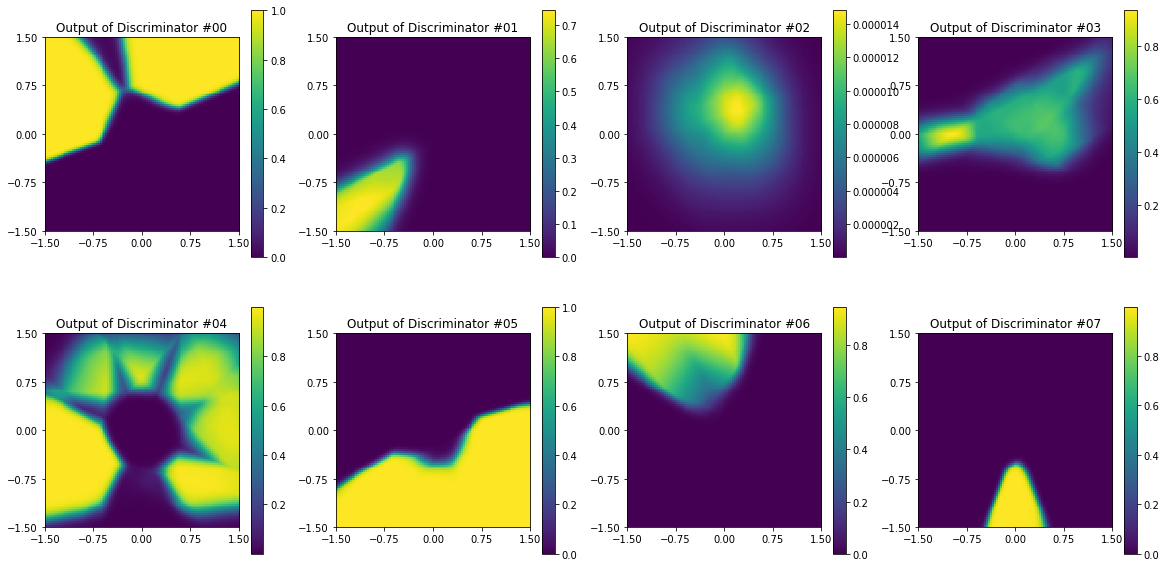}
    }\\
    \subfloat[]{
        \includegraphics[width=.85\linewidth]{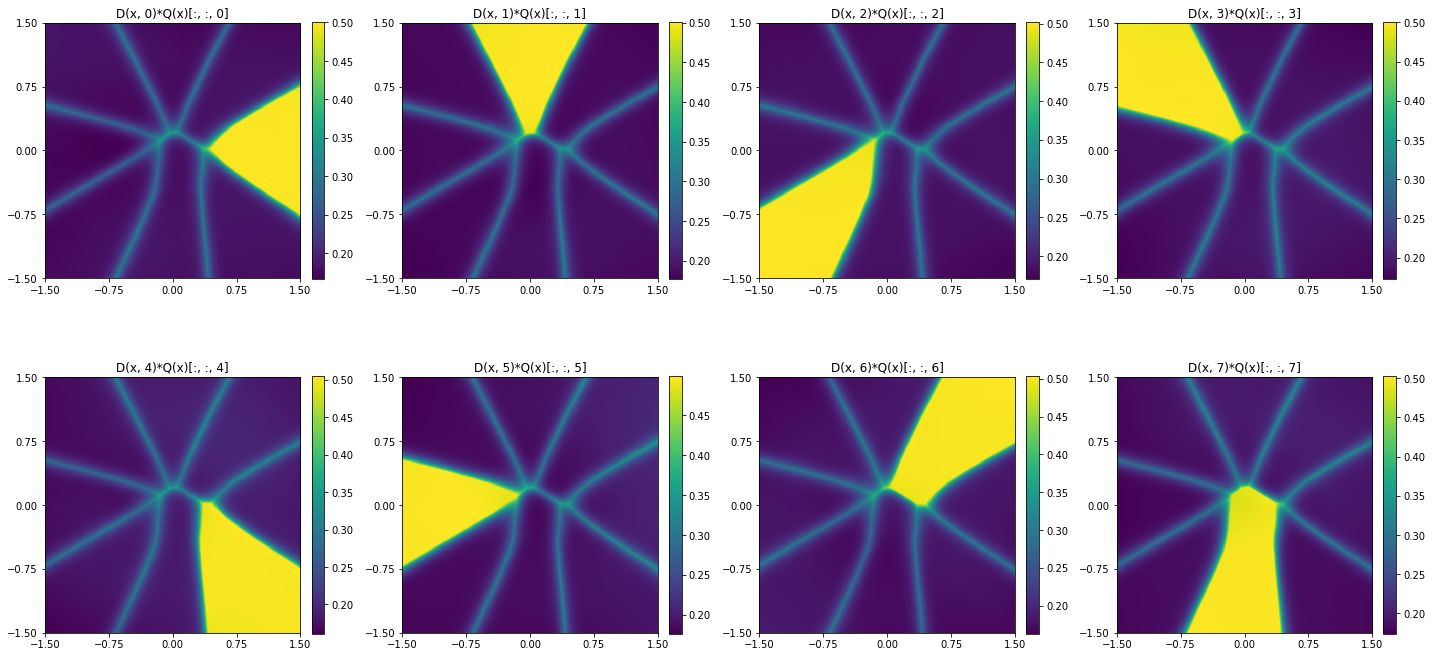}
    }
    
    \caption{Heat map of discriminator scores $\in [0, 1]$ ($0$ signifies fake while $1$ means real) when evaluated for every data-point pair lying in $\left[-1.5, 1.5\right]^2$ (corresponding to Figure~\ref{fig:8GMM}).
    (a) GMAN: it is clear that discriminator \#04 already covers the majority of the modes while the other discriminators give high scores for obviously fake samples (\#03 and \#05). 
    (b) \DPN{}: here we multiplied the discriminator scores with the probability of each point being assigned to that discriminator (obtained from classifier $Q$). 
    Although the capacity of each discriminator in \DPN{} is identical to the discriminators in the GMAN experiment, the \DPN{} framework reduces the complexity of each discriminator's task by making it work only on a different identifiable mode.
    }
    \label{fig:8GMM-disc}
    }
\end{figure*}

As discussed in the section~\ref{sub:obs}, here we show qualitative results with $N=M=8$ (where $N$ is the number of discriminators, $M$ is the number of modes) whose corresponding quantitative results are mentioned in the Table~\ref{tab:2d}. We illustrate our findings in Figure~\ref{fig:8GMM} and Figure~\ref{fig:8GMM-disc}.

We argue that GMAN is not able to utilize multiple discriminators in this experiment setup and that most of the learning is done by just a few discriminators rather than their effective ensemble (see Appendix~\ref{app:2d}).

In (4.1.2), under Error Analysis paragraph we claimed that GMAN fails to utilize multiple discriminators to their full potential. In Figure~\ref{fig:2d} we already have visual proof: the gradient field of the first two discriminators (top row, red and blue) are almost identical to each other, while the gradient of the third network (top row, green) is pointing towards a completely different mode (lower left) in its non-adjacent area, while around this distant mode the magnitude of the gradient is relatively small.

\section{CIFAR-10}
\label{app:cifar}
\begin{figure*}[!ht]
    \centering{
    \subfloat[Ship]{
        \includegraphics[width=.3\linewidth]{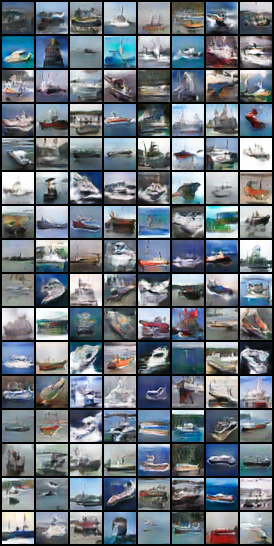}
    }
    \subfloat[Horse]{
        \includegraphics[width=.3\linewidth]{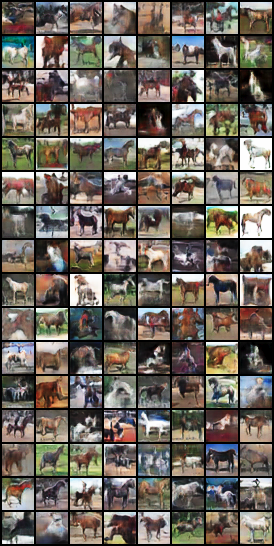}
    }
    \subfloat[Car]{
        \includegraphics[width=.3\linewidth]{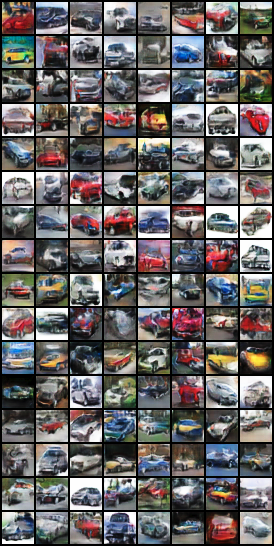}
    }\\
    \subfloat[Airplane]{
        \includegraphics[width=.3\linewidth]{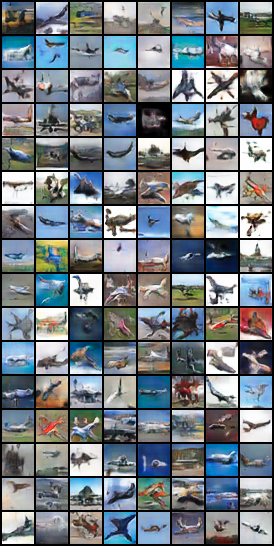}
    }
    \subfloat[Cat]{
        \includegraphics[width=.3\linewidth]{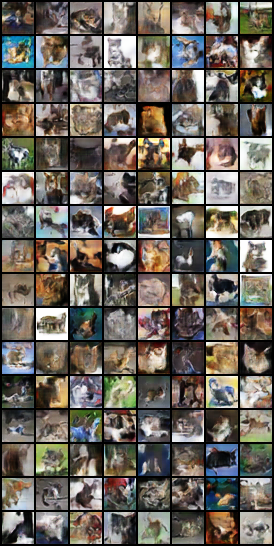}
    }
    \subfloat[Deer]{
        \includegraphics[width=.3\linewidth]{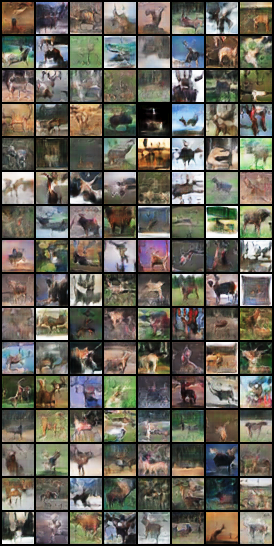}
    }
    \caption{$32\times 32$ CIFAR-10 samples drawn from \DPN{} trained with $N=5$ discriminators for 700k iterations. The generations in each subfigure correspond to $y=1,\dots , 6$ respectively.
    We call the attention to the various different details learnt for each class which can cause mode collapse in other GAN variants.
    }
    \label{fig:cifar10_more}
    }
\end{figure*}

\begin{figure*}
    \centering{
    \subfloat[Dog]{
        \includegraphics[width=.3\linewidth]{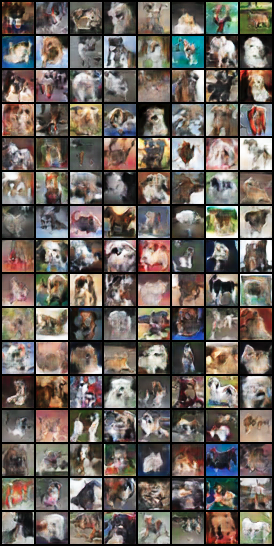}
    }
    \subfloat[Frog]{
        \includegraphics[width=.3\linewidth]{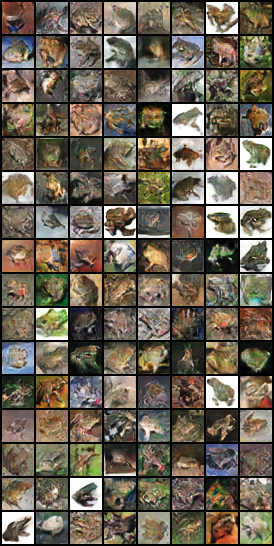}
    }\\
    \subfloat[Bird]{
        \includegraphics[width=.3\linewidth]{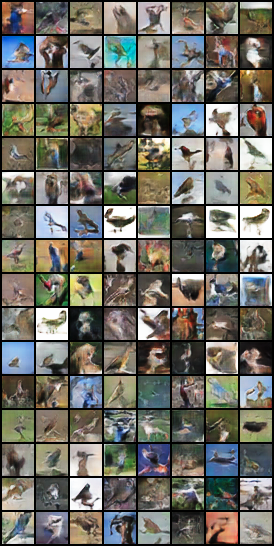}
    }
    \subfloat[Truck]{
        \includegraphics[width=.3\linewidth]{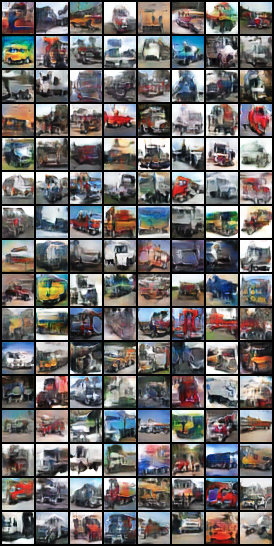}
    }
    \caption{$32\times 32$ CIFAR-10 samples drawn from \DPN{} trained with $N=5$ discriminators for 700k iterations. The generations in each subfigure correspond to $y=7,8,9$ and $10$ respectively.
    We call the attention to the various different details learnt for each class which can cause mode collapse in other GAN variants.
    }
    \label{fig:cifar10_more1}
    }
\end{figure*}

As discussed in the Section~\ref{subsubsec:qualImg}, here we present some more results obtained for each of the $10$ classes of CIFAR-10 in Figure~\ref{fig:cifar10_more} and Figure~\ref{fig:cifar10_more1}. 

\section{Implementation details}
\label{app:imp_det}
Here we present the way we structured our experiments and the details about the network architecture we used in the experiments.
    \subsection{Synthetic low dimensional distributions}
    First, we reproduced the 1D setting in \cite{ghosh2017multi} with 5 modes at $[10, 20, 60, 80, 110]$ and standard deviations $[3, 3, 2, 2, 1]$ respectively and we compare to the numbers reported in that paper in Table~\ref{tab:1d-var}. 
    Second, we compared \DPN{} directly to GMAN~\cite{durugkar2016generative} qualitatively in Figure~\ref{fig:2d} and quantitatively in Table~\ref{tab:2d} using a circular 2D GMM distribution with 3 and 8 modes respecitvely, around the unit circle.
    To illustrate the advantage of \DPN{} over GMAN~\cite{durugkar2016generative} we plotted the gradient field to visualize the benefit of using multiple discriminators. The gradient field of this setup can be seen in Figure~\ref{fig:2d}. 
    To get quantitative results, we estimated the probability density distribution using a histogram with 1400 bins over the real and generated samples and computed the Chi-square and KL-divergence between the two histograms.
        \paragraph{Comparing against other GAN variants}
        When comparing against other GAN variants, we run the 1D experiments using a fixed set of 200,000 samples from the real distribution and generate 65,536 elements from each model. 
        
        Since \DPN{} is directly designed to separate different modes, we outperform all the other methods as shown in Table \ref{tab:1d-var}. 
        
        In our case, we sample the code vectors for the generator from a categorical distribution with uniform probability. For the best results, we use 5 discriminators in both GMAN \cite{durugkar2016generative} and \DPN{}. For both, we train 3 instances and select the best score from each of them.

        \paragraph{Benchmarking the number of discriminators} 
        In 1D for better non-parametric probability density estimation, we increased the number of generated samples from 65,536 to 1,000,000 samples as done in \cite{ghosh2017multi}.
        For more reliable results on the implied mechanism of both approaches, we run the training 20 times for each algorithm with number of discriminators $N={2,\ldots,8}$, totaling 320 training. As in the previous experiment, we chose the best results from each run. 
        
        \paragraph{2D experiments}
        In 2D, for both variants we experiment with $M=N=8$ (where $N$ is the number of discriminators, $M$ is the number of Gaussians we used in the mixture) for quantitative results, listed in Table \ref{tab:2d} and $M=N=3$ setting for qualitative results, illustrated in Figure \ref{fig:2d}.
        For each experiment we use a fixed set of 1,000,000 samples and take 5 run per each algorithm, then report the best run.
        We took effort to make sure that the comparison was fair, and used the same set of parameters as it was done in the 1D experiments.
        
    \subsection{Image generation}
        For both the generator and discriminator we use ResNet-architectures~\cite{he2016deep}, with 18 layers each in the CIFAR-10 experiments, and 26 layers each in the CelebA experiments.
        As was done in \cite{mescheder2018convergence} we multiply the output of the ResNet blocks with 0.1, use 256-dimensional unit Gaussian distribution.
        For categorical conditional image generation we use an embedding network that projects category indices to 256 dimensional label vector, normalized to the unit sphere.
        In the case of conditional image generation the classifier $Q$ is trained on code vectors, so it is constrained to learn the original class labels.
        We embed the code vector similar to the ground truth labels in this setting for CIFAR-10.
        We use Leaky-RELU nonlinearities everywhere, without BatchNorm.
        
        Following the considerations in \cite{mescheder2018convergence} for optimizing parameters of $Q$, $D$, $G$ we use the RMSProp with $\alpha=0.99$, $\epsilon=10^{-8}$, and initial learning rate of $10^{-3}$.
        We use a batch size of 64, and train the algorithm for 700,000 and 400,000 iterations for CIFAR-10 and CelebA tasks respectively.
        Similar to work that provided state of the art results on image generation tasks \cite{karras2017progressive, mescheder2018convergence} for visualizing the generator's progress we use an exponential moving average of the parameters of $G$ with decay 0.999.

